\documentclass{article}

\usepackage{iclr2027_conference,times}

\usepackage{graphicx}
\usepackage{subcaption}

\usepackage[utf8]{inputenc} % allow utf-8 input
\usepackage[T1]{fontenc}    % use 8-bit T1 fonts
\usepackage{hyperref}       % hyperlinks
\usepackage{url}            % simple URL typesetting
\usepackage{booktabs}       % professional-quality tables
\usepackage{amsfonts}       % blackboard math symbols
\usepackage{nicefrac}       % compact symbols for 1/2, etc.
\usepackage{microtype}      % microtypography
\usepackage{xcolor}         % colors
\usepackage{multirow}
\usepackage{bbm}
\usepackage{amsmath}
\usepackage{amsthm}
\usepackage{algorithm}
\usepackage{algorithmic}

\usepackage[table]{xcolor}
\definecolor{rowlightblue}{RGB}{225,242,255}
\definecolor{rowlightgrey}{RGB}{240,240,240}
\newcommand{\greycell}{\cellcolor{rowlightgrey}}
\newcommand{\bluecell}{\cellcolor{rowlightblue}}
\usepackage{caption}
\usepackage{placeins}  % FloatBarrier
\usepackage{adjustbox}

\newcommand{\dd}[1]{{\scriptsize\textcolor{gray}{($-$#1)}}}

\makeatletter
\newif\ifinappendix
\AddToHook{cmd/subsection/before}{\ifinappendix \FloatBarrier \fi}
\AddToHook{cmd/subsubsection/before}{\ifinappendix \FloatBarrier \fi}
\makeatother

\title{LowRAM: Low-Rank Keys and Amortised Values for Composable KV Cache Compression at Test-Time}

\author{%
Teresa R. Delgado$^{1,*}$, Hugo Hazard$^{2,*}$, Martin Benfeghoul$^{1,2,*}$, \\
\textbf{Jun Wang$^{2}$, Zafeirios Fountas$^{1}$} \\
$^{1}$Huawei, Noah's Ark Lab, London, UK \\
$^{2}$AI Centre, Department of Computer Science, University College London, London, UK \\
\texttt{teresa.delgado@h-partners.com} \\
\texttt{zafeirios.fountas@huawei.com} \\
}

\begin{document}

\iclrfinalcopy % removes line numbers and shows author names

\maketitle
\lhead{Preprint} % overrides the ICLR running header
{\renewcommand{\thefootnote}{}\footnotetext{$^{*}$Equal contribution.}}

\begin{abstract}
Autoregressive decoding relies on the KV cache, whose memory footprint grows linearly with context length and caps batch size and throughput. Existing post-training methods exploit redundancy across tokens, layers, and feature dimensions, or lower numerical precision. We identify a complementary axis: within each attention head, values are largely predictable from their corresponding keys. We exploit this with value cache amortisation, which fits a lightweight per-head predictor at test time and stores residuals only for the tokens it reconstructs worst, giving $4\times$ near-lossless value compression. The predictor can be initialised by meta-learning for substantially reduced prefill overhead, or fitted from scratch in settings that allow a one-off cost, such as multi-query inference. We pair value amortisation with cross-layer low-rank key compression to produce LowRAM, an asymmetric compression pipeline that stays close to lossless across three model families. By reconstructing only a selected subset of KV pairs at each decoding step, LowRAM can keep decoding overhead minimal and achieve up to $5\times$ higher peak decode throughput at $4\times$ compression. LowRAM further composes with quantisation and eviction, reaching $26.9\times$ compression at $94.5\%$ of baseline performance on long contexts. These results show that treating keys and values as structurally distinct enables efficient, aggressive, and composable KV cache compression.
\end{abstract}

\section{Introduction}
Transformers~\citep{vaswani2017attention} have become the dominant architecture for language modelling, but their deployment at long context lengths is fundamentally constrained by the compute cost of autoregressive decoding. Modern inference stacks address this via a KV cache: the key and value projections of each preceding token are computed once and reused at every subsequent decoding step. This turns decoding from a quadratic to a linear operation in sequence length, but shifts the bottleneck from compute to memory: the cache grows linearly with context length and batch size, and at long contexts comes to dominate the model's activation footprint.

\begin{figure}[htbp]
\centering
\small
\begin{subfigure}[t]{0.48\linewidth}
    \centering
    \includegraphics[width=\linewidth]
        {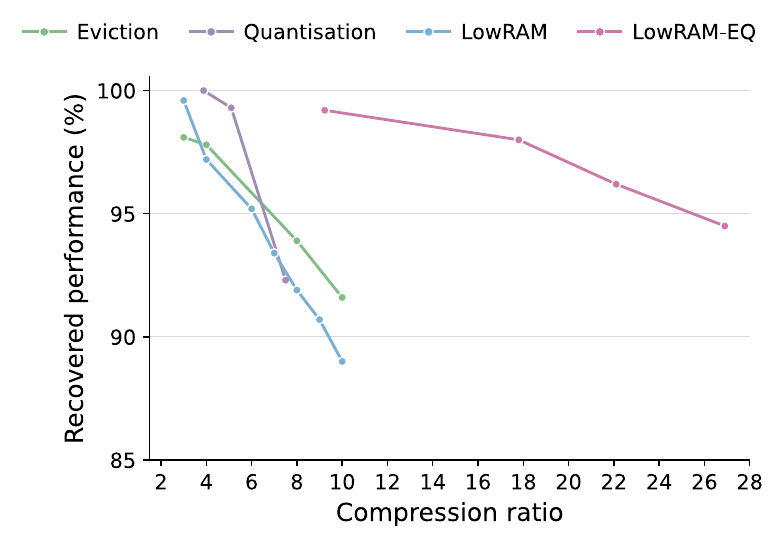}
    \label{fig:pareto_frontier}
\end{subfigure}
\hfill
\begin{subfigure}[t]{0.48\linewidth}
    \centering
    \raisebox{8pt}[0pt][0pt]{%
        \includegraphics[width=\linewidth]
            {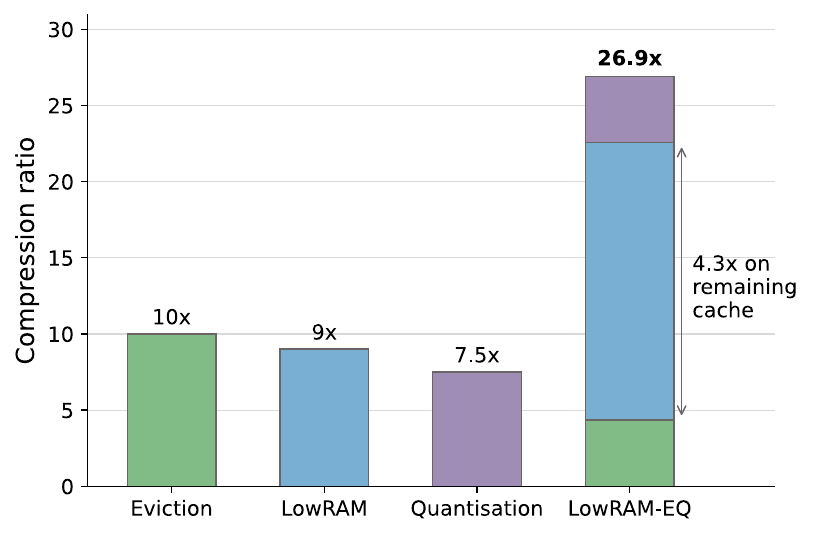}%
    }
    \label{fig:extreme_compression}
\end{subfigure}
\caption{
Mistral-7B compression on LongBench contexts longer than 10K tokens across different KV cache compression methods: eviction (adapted from~\cite{yang2025attentionpredictor}), quantisation~\citep{zandieh2026turboquant}, LowRAM (our proposed method), and LowRAM-EQ (LowRAM combined with eviction and quantisation). \textbf{Left}: recovered performance against compression ratio. \textbf{Right}: maximum evaluated compression retaining at least 90\% of baseline performance. LowRAM-EQ combines LowRAM, eviction, and quantisation. See Appendix~\ref{app:comp_ratio} for a description of how compression is computed.
}
\label{fig:mistral_compression}
\end{figure}

A wide range of architectural responses to this bottleneck have been proposed, including linear attention and state-space alternatives that replace the cache with a fixed-size recurrent hidden state~\citep{pmlr-v119-katharopoulos20a, gu2023mamba, dao2024transformers}, and sparse, sliding-window, grouped-query, and multi-head latent attention variants that reduce per-token cache size architecturally~\citep{ainslie2023gqa, deepseekai2024deepseekv2strongeconomicalefficient}. However, these approaches require pretraining or substantial finetuning, and most open-weight models still rely on grouped-query attention with a full KV cache \citep{grattafiori2024llama, jiang2023mistral7b, gemmateam2024gemma2improvingopen, yang2025qwen3technicalreport}. This has motivated a parallel line of \emph{post-training} KV cache compression methods operating on frozen pretrained models. Existing methods fall into three families: eviction, which discards unimportant entries~\citep{xiao2024efficient, zhang2023ho, li2024snapkv, cai2025pyramidkv}; quantisation, which lowers the precision of stored entries~\citep{liu2024kivi, hooper2024kvquant, dong2024qaq}; and low-rank decomposition, which approximates KV matrices with smaller, factorised representations~\citep{yu2024eckvh, chang2024palu, saxena2024eigen}. These families are largely orthogonal, exploiting redundancies along different axes of the cache. This leaves the redundancy between keys and values largely unexplored. Keys and values are computed from the same representation, and although they serve different functions, their shared origin induces a predictable relationship between them. We exploit this redundancy by using the keys as a source from which the value cache can be recovered.

\begin{figure}
    \centering
    \small
    \includegraphics[width=1\linewidth]{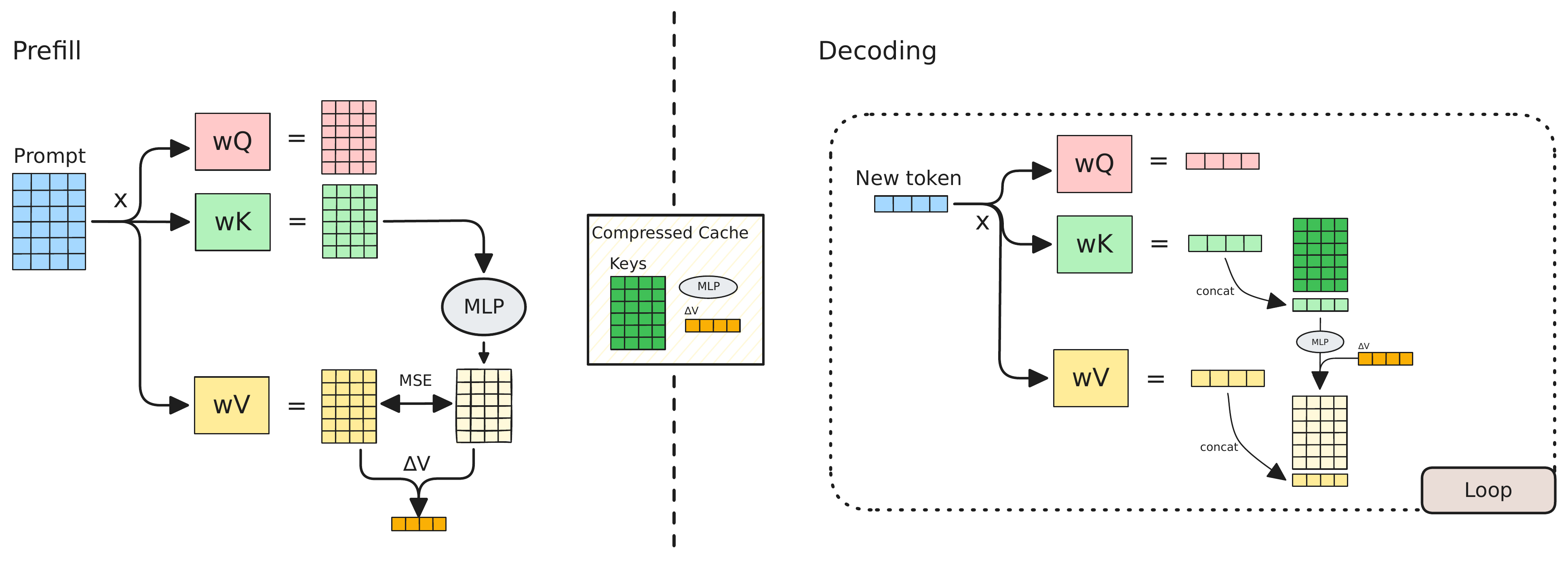}
    \caption{At inference, a per-head MLP learns to reconstruct values from keys, then is cached for decoding alongside residuals for the worst-reconstructed values.}
    \label{fig:architecture}
\end{figure}

We first compare several predictive relationships within and across layers under a matched memory budget and find that reconstructing current-layer values from current-layer keys best preserves downstream performance. Based on this finding, we introduce value cache amortisation: during prefill, we fit a lightweight per-head network to map cached keys to their associated values. We then rank tokens by reconstruction error and store residual corrections only for the least accurately predicted values. The resulting value cache consists of the network parameters and a sparse set of residuals. During decoding, values are recovered by applying the network to the keys and adding the stored corrections, as illustrated in Figure~\ref{fig:architecture}. 

Amortising the values still requires keys to be retained, both for attention and as inputs to the learned map. A complete compression scheme must therefore reduce the key cache as well. Our source-target analysis suggests that applying the same predictive strategy to the keys leads to less downstream performance than for values. Instead, for keys, prior work points to low-rank decomposition: their energy is concentrated in a relatively small number of feature directions, producing a short spectral tail and correspondingly small approximation error at reduced rank~\citep{saxena2024eigen}. These findings motivate an asymmetric design: values are reconstructed from keys, while the keys themselves are compressed according to their low-rank structure. LowRAM realises this design by pairing value cache amortisation with xKV~\citep{chang2026xkv}, which shares a low-rank basis across layers within each group to exploit alignment between their dominant key-cache subspaces.

\paragraph{Contributions.}
\begin{itemize}
    \item We identify within-head key-to-value predictability as a previously
    underexplored source of KV-cache redundancy. Building on this observation,
    we introduce value cache amortisation, which replaces most stored values
    with a context-adapted predictor and retains residuals only for the
    hardest-to-predict tokens.

    \item We propose LowRAM, an asymmetric pipeline that combines amortised
    values with cross-layer low-rank key compression. LowRAM remains close to
    lossless at moderate compression across three model families and composes
    with quantisation and eviction, reaching \(26.9\times\) compression while
    retaining \(94.5\%\) of baseline performance.

    \item We reduce the computational overhead of LowRAM using a
    meta-learned initialisation and selective reconstruction. The former dramatically reduces prefill overhead introduced by amortisation, while the latter achieves up to \(5\times\) higher peak decode throughput. 
\end{itemize}

\section{Related Works}
\paragraph{Low-Rank Decomposition of KVs.}\label{sec:related_works__low_rank}
Training-free SVD-based KV compression largely targets per-layer redundancy, via activation-aware or grouped SVD on the KV projection weights~\citep{yuan2025asvdactivationawaresingularvalue,yu2024eckvh,zhang2024lorc,chang2024palu,li2025recalkv}, projecting attention into an offline-fitted eigen-subspace~\citep{saxena2024eigen}, or direct channel pruning~\citep{xu2024think}. Complementary learned-projection methods~\citep{lin2024matryoshkakv,kvlatent2025,stief2026,deltakv2026,echokv2026,kqsvd2026,ojakv2025,lecomte2026} reach similar ratios with trained or online-adapted projectors. xKV~\citep{chang2026xkv} instead exploits cross-layer redundancy, sharing a single SVD basis across contiguous layer groups since adjacent layers share dominant singular vectors.

\paragraph{Eviction.}\label{sec:related_works__eviction}
Eviction reduces KV cache memory by discarding unimportant tokens. Early methods keep attention sinks and a recent window~\citep{xiao2024efficient}, or a mix of recent and high-attention tokens~\citep{zhang2023ho}. Later work selects tokens using an observation window~\citep{li2024snapkv}, or allocates budgets per layer~\citep{cai2025pyramidkv} or per head~\citep{ge2024model,feng2025adakv,tang2025razorattention}. Query-agnostic methods instead score tokens by how well they support reconstructing the context~\citep{kim2025kvzip,jegou2026kvzap}.

\paragraph{Quantisation.}\label{sec:related_works__quantisation}
Quantisation compresses the KV cache by lowering the precision of stored keys and values. However, these exhibit channel-wise outliers \citep{dettmers2022gptint,xiao2023smoothquant} that dominate a uniform quantisation grid, so naive quantisation degrades sharply below $4$ bits \citep{liu2024kivi,hooper2024kvquant}. Sub-$4$-bit methods therefore target these outliers by quantising keys channel-wise and values token-wise~\citep{liu2024kivi}, allocating bit-widths per token~\citep{yang2024mikv,he2024zipcache,dong2024qaq} or per layer~\citep{li2025kvtuner}, inverting positional embeddings before quantisation~\citep{hooper2024kvquant}, or removing outliers with random rotations~\citep{ashkboos2024quarot,han2026polarquant,zandieh2026turboquant}. AQUA-KV \citep{shutova2025cache} instead quantises a prediction residual, using linear adapters that exploit both inter-layer and intra-layer dependencies: keys are predicted from previous-layer keys, while values are predicted jointly from previous-layer values and current-layer keys. Our amortised value cache uses context-adapted prediction of values from keys within each attention head, storing residuals only for tokens that are hard to reconstruct.

\section{Methods}
\label{sec:methods}
We propose LowRAM, an asymmetric compression pipeline that treats the key and value caches as structurally distinct objects. For keys, we adopt xKV~\citep{chang2026xkv}, which exploits the low-rank cross-layer redundancy of key states. For values, we introduce amortisation, a learned associative memory that predicts each token's value vector directly from its key, fit once per context, and retains residuals only for the values the predictor reconstructs poorly. Prior to both decomposition and prediction, we invert RoPE, following evidence that pre-RoPE representations are more amenable to compression~\citep{hooper2024kvquant, su2025rotatekvaccuraterobust2bit, sun2025shadowkv}, with the rotation reapplied at attention time. Both predictor fitting and value reconstruction use the pre-RoPE keys reconstructed by xKV. During decoding, generated key-values are stored exactly in the cache.

\subsection{Identifying the Predictive Structure of the KV Cache}\label{sec:source_target}
Prior work has explored predictive compression of the KV cache using inter-layer and intra-layer dependencies ~\citep{shutova2025cache} or across tokens~\citep{deltakv2026}.Here, we focus on learning a mapping from keys to values within individual attention heads. To determine which representation is best suited to predictive compression within our framework, we perform a source-target analysis over the KV cache. For each layer, we train the same predictor to reconstruct either the current-layer keys or values while varying the available source representation. Specifically, we consider previous-layer keys ($K_{L-1}$), previous-layer values ($V_{L-1}$), their concatenation ($KV_{L-1}$), current-layer values ($V_L$), and current-layer keys ($K_L$). To match our proposed compression scheme, each predictor is paired with a residual store, with the residual budget fixed across all configurations. All configurations are evaluated under the same $3\times$ compression budget using both reconstruction error and recovered LongBench performance.

\begin{figure}
    \centering
    \small
    \includegraphics[width=1.\linewidth]{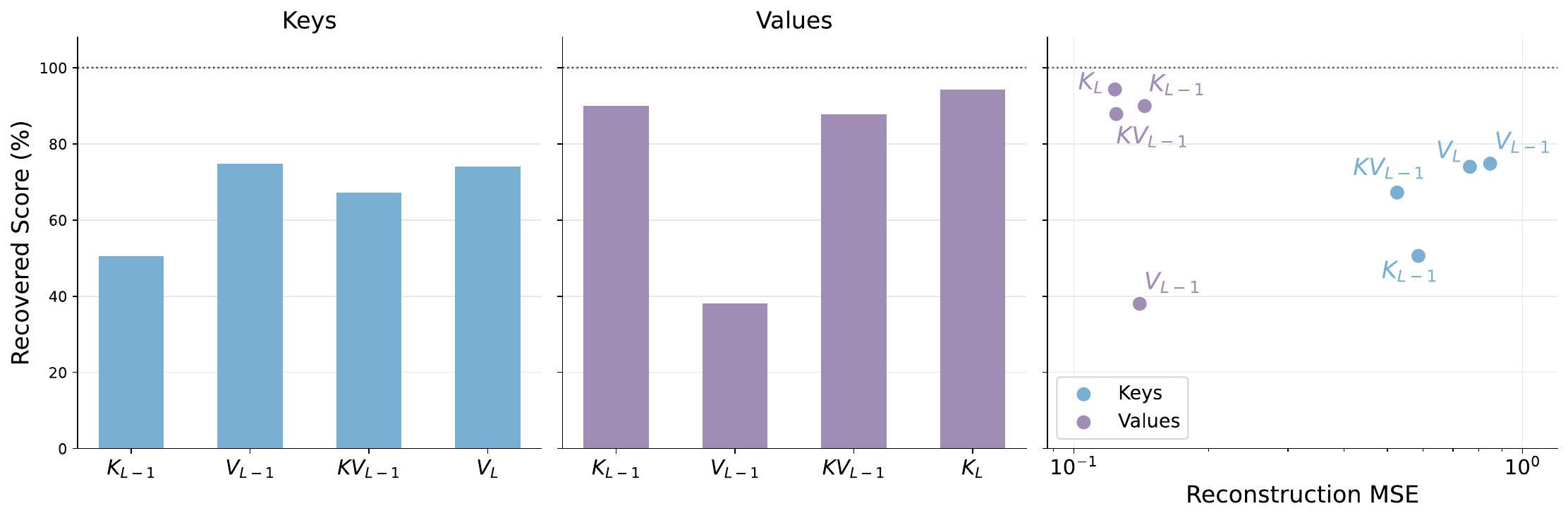}
    \caption{Source-target analysis for predictive KV cache compression at a fixed $3\times$ compression ratio for Mistral-7B. The left and centre panels show recovered LongBench performance when reconstructing current-layer keys $K_L$ and values $V_L$, respectively, from alternative current- or previous-layer representations. The dotted line indicates uncompressed baseline performance. The right panel plots recovered performance against reconstruction mean-squared error for each source-target pair, showing that lower reconstruction error does not necessarily correspond to better downstream performance.}
    \label{fig:analysis}
\end{figure}

Figure \ref{fig:analysis} shows that reconstructing values from current-layer keys, $K_L \rightarrow V_L$, achieves the highest recovered downstream performance among the configurations evaluated. This source-target pair therefore forms the basis of our proposed method. Our comparison revisits the probing configurations examined by AQUA-KV~\citep{shutova2025cache} (Figure 2), evaluating them within our predictor and residual-storage framework using both reconstruction error and recovered LongBench performance. Under this setting, lower reconstruction error does not necessarily imply better preserved downstream performance.

Two further observations are worth noting. First, value reconstruction generally recovers more downstream performance than key reconstruction. Second, reconstruction error correlates only weakly with downstream performance: source–target pairs with similar reconstruction errors can exhibit significantly different recovered performance, and the differing scales of keys and values make their absolute error difficult to compare directly. We therefore base our design decisions primarily on recovered downstream performance rather than reconstruction error.

\subsection{Value Cache Amortisation}
Having established $K_L \rightarrow V_L$ as the preferred prediction direction, we model this relationship with a lightweight parametric function. Let $K\in \mathbb{R}^{T \times d_k}$ and $V\in \mathbb{R}^{T \times d_v}$ denote the key and value matrices for a single attention head over a sequence of length $T$. We learn a mapping
$f_\theta : \mathbb{R}^{d_k} \rightarrow \mathbb{R}^{d_v}$ applied row-wise to the key cache, independently per layer $\ell$ and KV head $h$. Concretely, $f_\theta$ is a two-layer MLP with a learned linear residual branch,
\begin{equation}
    f_{\theta_{\ell,h}}(k)
    =
    \sigma\!\left(
        kW^{(1)}_{\ell,h}+b^{(1)}_{\ell,h}
    \right)
    W^{(2)}_{\ell,h}
    + b^{(2)}_{\ell,h}
    + kW^{\mathrm{lin}}_{\ell,h},
\label{eq:predictor_arch}
\end{equation}
where $\sigma$ is a ReLU nonlinearity and $W^{\mathrm{lin}}_{\ell,h}$ is initialised from the Slim Attention prior introduced below; Appendix~\ref{app:reconstruction_training} gives the full architecture and layer-wise prefill-fitting procedure. During prefill, the parameters $\theta$ are adapted to the current context by minimising the value reconstruction loss $\mathcal{L}(\theta) = \frac{1}{T}\|f_\theta(K) - V\|_F^2$.

At decoding time, the value cache is reconstructed from the retained keys as $\hat{V} = f_\theta(K)$. This prediction captures much of the value cache, but not all tokens are reconstructed equally well. Therefore, we store residual corrections only for the tokens with the largest prediction errors: for each token $i$, the reconstruction error is $e_i = \left\lVert f_\theta(k_i) - v_i \right\rVert_2^2$.

Let $p \in (0,1]$ denote the fraction of residuals retained. We select the top-$p$ fraction of tokens according to $e_i$ and define a binary gate 
\begin{equation}
g_i =
\mathbbm{1}
\left[
\mathrm{rank}(e_i) > (1-p)T
\right]
\label{eq:gate}
\end{equation}
where larger rank corresponds to larger reconstruction error. Writing $\Delta v_i = v_i - f_\theta(k_i)$ for the per-token residual, the reconstructed value cache is, in matrix form, 

\begin{equation}
\tilde{V}
=
f_\theta(K) + G \odot \Delta V
\label{eq:v_tilde}
\end{equation}

which replaces the original value cache in standard attention,
$O = \mathrm{softmax} \left(QK^\top / \sqrt{d_k} \right) \tilde{V}$. Algorithm~\ref{alg:pem} in Appendix~\ref{app:reconstruction_training} summarises this procedure end to end.

The parameter $p$ controls the memory-fidelity trade-off. Smaller $p$ stores fewer residuals and hence increases compression, while larger $p$ retains more of the original value cache. The total memory footprint consists of two terms: a sequence-independent term from the per-head MLP parameters and the linear residual $W_{\mathrm{lin}}$, and a sequence-dependent term from the retained residuals and their indices. Figure~\ref{fig:memory_footprint} shows that the fixed parameter cost is amortised after the first few thousand tokens, after which memory grows at a fraction of the uncompressed value cache slope.

\paragraph{Initialisation from Slim Attention. }\label{sec:slim_init}
Following Slim Attention~\citep{graef2025slimattentioncutcontext}, we initialise the residual branch $W_{\text{lin}}$ from the key and value projection matrices. When the key projection $W_K$ is square and invertible, values can be reconstructed exactly from keys via $W_K^{-1}W_V$; in GQA models $W_K$ is rectangular, so we instead use the Moore-Penrose pseudoinverse, $W_{\text{lin}} = W_K^{+}W_V$. This gives the best linear approximation to the key-to-value mapping the model weights can offer, and a strong starting point for training, on top of which the MLP learns a non-linear correction that captures value components the Slim map cannot represent in GQA models.

\paragraph{Warm-starting the Amortised Value Cache MLP. }
Adapting $f_\theta$ from scratch on each context wastes gradient steps relearning shared structure. We instead split the work in two: a one-off offline stage that learns the general key-to-value mapping (one pass over 32 batches of 4096-token FineWeb-Edu sequences, 25 gradient steps, under five minutes on one GPU, run once per model), and an online stage that overfits this mapping to the current context. At inference, these weights initialise each layer's MLP before per-context adaptation, reducing the steps needed to reach a useful approximation.

\subsection{Efficient Adaptation and Selective Reconstruction}
\label{sec:efficiency_methods}

\paragraph{Meta-learned initialisation.}
Even with warm-starting, fitting the value predictor requires many per-context gradient updates, making it costly when a context is queried only once. We therefore meta-learn an initialisation \(\phi\) using First-Order MAML~\citealp{nichol2018first}. Each sequence defines a meta-learning task consisting of its observed KV pairs, and the inner loop adapts the per-head predictor for \(K\) gradient updates. At inference, the predictor is initialised from the learned parameters \(\phi^\star\) and adapted to the new context using the same \(K\)-update budget. Appendix~\ref{app:meta_learning} gives the complete training procedure.

\paragraph{Selective reconstruction.}
Reconstructing the complete cache at every decoding step can offset the computational benefit of compression. Following landmark-guided retrieval~\citep{sun2025shadowkv}, we partition the prefix into chunks represented by mean-key landmarks, retain outlier chunks and a local window exactly, and reconstruct only the top-\(k\) chunks selected by the current query. A persistent buffer reuses chunks selected across consecutive decoding steps. We additionally replace the exact key cache SVD with a randomised Cholesky-QR decomposition, giving a $4\times$ speedup with no loss in downstream quality~\citep{chang2026xkv}. Appendix~\ref{app:selective_reconstruction} describes the retrieval configuration, kernels, and memory accounting.

\section{Results}
\subsection{Experimental Setup}\label{sec:experimental_setup}
\paragraph{Benchmarks.} We evaluate on RULER~\citep{hsieh2024ruler} and LongBench~\citep{bai-etal-2024-longbench}, and additionally report LongBench filtered to sequences of at least 10K tokens. For this subset we exclude tasks with fewer than 20 samples meeting the length requirement. Appendix~\ref{appdx:benchmarks} gives per-task statistics. All results are reported as recovered performance relative to the uncompressed baseline. We evaluate across three model families: Mistral-7B-Instruct-v0.3~\citep{jiang2023mistral7b}, Llama-3.1-8B-Instruct~\citep{grattafiori2024llama}, and Qwen3.5-4B~\citep{qwen3.5}. Qwen3.5-4B is a hybrid architecture that interleaves full attention with linear attention layers; only the full attention layers maintain a conventional KV cache, so LowRAM is applied to that subset of layers alone, and the compression ratios we report for Qwen3.5-4B are relative to that subset's cache rather than the whole model's. 

\paragraph{Eviction.}
When combined with LowRAM, we apply eviction before fitting the map, discarding tokens unlikely to be attended to so the MLP need only reconstruct and store residuals for the tokens it retains. Importance is estimated with a small CNN predictor adapted from~\citet{yang2025attentionpredictor}: during prefill, a pre-hook on each attention layer scores every token by predicted future attention (averaged across heads), and we retain the top-$\rho T$ positions by this score, with attention sinks (first 8 tokens) and a local window (last 64 tokens) kept unconditionally.

\paragraph{Quantisation.}
We quantise with TurboQuant \citep{zandieh2026turboquant}, omitting the 1-bit QJL \citep{zandieh2025qjl} error-correction step. Although QJL debiases the inner product estimations, we found that its variance is amplified by Softmax, thus degrading both attention quality and downstream accuracy. For the key cache, xKV factorises the layer-grouped cache as $K_l \approx AB_l$, where $A$ holds the shared low-rank basis and $B_l$ is layer-specific; since only $A$ scales with context length $T$, we quantise only $A$, which captures most of the memory reduction (Appendix~\ref{app:reconstruction_training} gives the crossover analysis for when quantising $B_l$ too becomes worthwhile). We use asymmetric precision, 4-bit for the key $A$ factor and 2-bit for value residuals.

\subsection{End-to-End Compression}
LowRAM retains near-baseline performance at moderate compression across all three model families (Table~\ref{tab:v_only_decomp_results}). Mistral-7B and Llama-3.1-8B retain \(97.9\)-\(99.4\%\) of baseline performance at 3\(\times\) compression. At 4\(\times\), recovery remains 96.0-99.6\% except for Llama-3.1-8B on RULER, where it falls to 93.9\%. We attribute the performance drop at this compression ratio to the reduced residual budget. As for Qwen3.5-4B, its RULER performance matches the uncompressed baseline, while LongBench recovery varies between 94.4\% and 97.8\%.

\begin{table}[h!]
    \centering
    \small
    \begin{tabular}{lcccccc}
        \toprule
            &  \multicolumn{3}{c}{RULER 4K (\%)} & \multicolumn{3}{c}{LongBench (\%)} \\
            \cmidrule(lr){2-4} \cmidrule(lr){5-7}
            Model & 2x & 3x & 4x & 2x & 3x & 4x \\
        \midrule
            \multirow{1}{*}{Mistral-7B}
                & 99.4 & 98.8 & 96.0 & 100.7 & 99.4 & 97.9  \\
        \midrule
            \multirow{1}{*}{Llama3.1-8B}
                & 99.6 & 97.9 & 93.9  & 98.7 & 98.5 & 99.6  \\
        \midrule
            \multirow{1}{*}{Qwen3.5-4B}
                & 100.0 & 100.0 & 100.1 & 97.8 & 95.0 & 94.4 \\
        \bottomrule
    \end{tabular}
    \caption{Recovered average performance of LowRAM on RULER 4K and LongBench. Scores are averaged across each benchmark and reported as a percentage of the corresponding uncompressed baseline. The compression ratio reported for Qwen3.5-4B concerns only the full attention layers.}
    \label{tab:v_only_decomp_results}
\end{table}

\subsection{Efficiency}
\label{sec:efficiency}
The results in Table~\ref{tab:v_only_decomp_results} were obtained by fitting the value-cache MLP with 150 gradient updates per context (300 in Table~\ref{tab:meta_selective_recon}, where performance has largely saturated, giving the strongest full-adaptation reference) and reconstructing the full cache at every decoding step, significantly increasing prefill and decoding cost, respectively. Section~\ref{sec:efficiency_methods} introduced one mechanism targeting each; here we measure their effect on latency, throughput and accuracy.

\begin{figure}[!ht]
    \centering
    \includegraphics[width=0.95\linewidth]{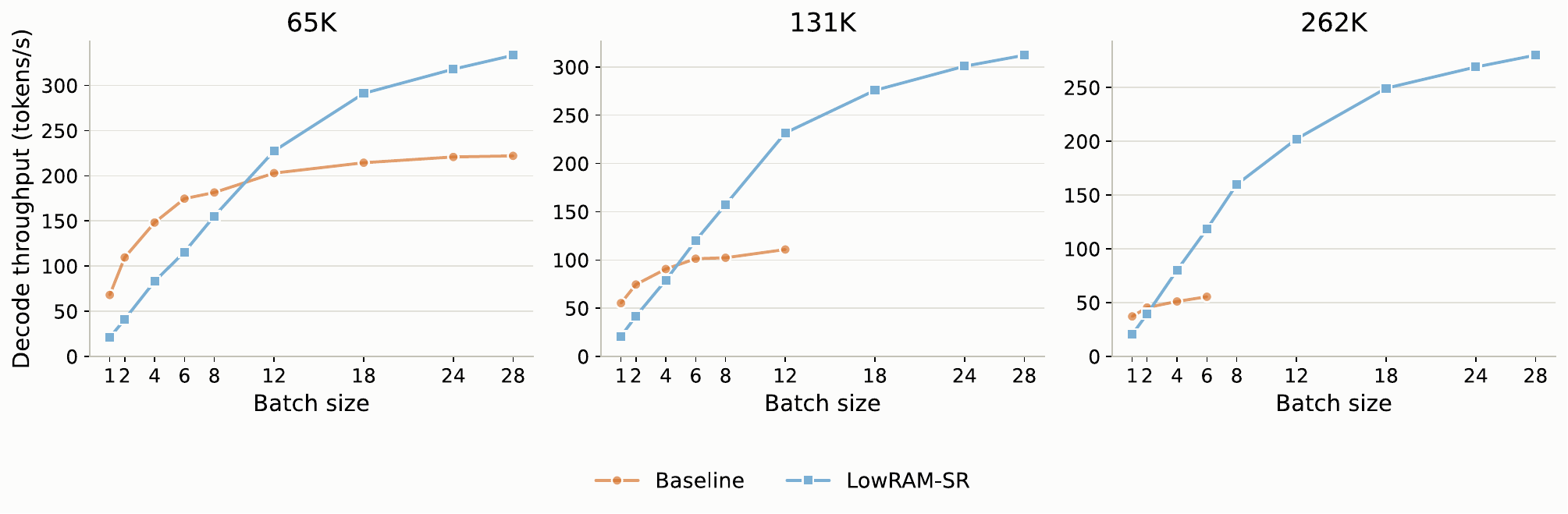}
    \caption{Decode throughput for Llama-3.1-8B at different batch sizes and sequence lengths at $4\times$ compression, comparing exact KV attention against LowRAM-SR (LowRAM with selective reconstruction). Appendix~\ref{app:selective_reconstruction} reports the full selective-reconstruction configuration.}
    \label{fig:throughput_main}
\end{figure}

Meta-learning the MLP initialisation reduces per-context adaptation to 2 gradient updates, lowering the time-to-first-token (TTFT) overhead of the amortised value cache to 1.3$\times$ that of the uncompressed baseline, down from 26.3$\times$ with 150 updates and 55.5$\times$ with 300 (Figure~\ref{fig:ttft_vs_epochs}). Figure~\ref{fig:throughput_main} compares the decode throughput of the baseline with that of LowRAM with selective reconstruction (LowRAM-SR) at 4$\times$ compression. The baseline is faster at small batch sizes, but it quickly becomes memory-bound and plateaus, while LowRAM-SR continues to scale. The crossover occurs earlier at longer contexts, and beyond it the gap widens until the baseline runs out of memory. LowRAM-SR supports a batch size of 28 at all three context lengths and, at 262K tokens, reaches roughly 5$\times$ the peak throughput of the baseline.

Table~\ref{tab:meta_selective_recon} reports the accuracy cost of each mechanism at $4\times$ compression. Selective reconstruction matches full reconstruction on LongBench ($>$10K) and stays within 1 point of it, or above it, at every RULER context length. The remaining difference comes from meta-initialisation, which costs 1.4-4.4 points on RULER and 2.4 on LongBench ($>$10K) relative to 300-update adaptation, in exchange for the TTFT reduction above. With both mechanisms enabled, LowRAM recovers over 92.4-96.8\% of baseline performance on RULER and 92.9\% on LongBench ($>$10K). Meta-initialisation is therefore the natural choice when a context is queried once. When the same context serves many queries, the cost of full adaptation is paid only once, and its extra accuracy comes at little amortised cost.

\begin{table}[t]
    \centering
    \small
    \begin{tabular}{cccccc}
        \toprule
        & & \multicolumn{3}{c}{RULER (\%)} & LongBench (\%)\\
        % \cmidrule(lr){3-6}
        \cmidrule(lr){3-5} \cmidrule(lr){6-6}
        SR & Meta-init & 16K & 32K & 64K & $>$10K\\
        \midrule
        \texttimes & \texttimes & 96.9 & 100.0 & 95.8 & 95.3 \\
        \checkmark & \texttimes & 97.7 & 99.0 & 96.8 & 95.3  \\
        \checkmark & \checkmark & 96.3 & 96.8 & 92.4 & 92.9 \\
        \bottomrule
    \end{tabular}
    \caption{Recovered LongBench ($>$10K) and RULER 16-64K performance of LowRAM on Llama-3.1-8B-Instruct at 4$\times$ compression, with and without selective reconstruction (SR) and a meta-learned initialisation (Meta-init). Rows without Meta-init use 300 gradient updates per context (vs. 150 in Table~\ref{tab:v_only_decomp_results}); Meta-init uses 2.}
    \label{tab:meta_selective_recon}
\end{table}

\subsection{Stacking with Eviction and Quantisation}

Table~\ref{tab:stacking_results_3x} evaluates LowRAM in combination with eviction and quantisation, and Table~\ref{tab:stacking_results} reports a full sweep in which both are applied on top of LowRAM at 2$\times$, 3$\times$ and 4$\times$ compression. Quantisation combines particularly well with our method, increasing both the overall compression ratio and recovered performance. For Mistral on LongBench, for example, adding quantisation to the 4$\times$ LowRAM cache raises compression to 6.4$\times$ and recovery from 97.9\% to 99.4\%. The benefit is largest where LowRAM alone degrades most, e.g.\ raising recovery for Llama on RULER at 4$\times$ from 93.9\% to 99.4\%. We attribute this to two factors. First, storing residuals at lower precision reduces the memory each one requires, so LowRAM can retain more residuals within the same budget. Second, in most configurations quantisation alone introduces very little error, so the gain from additional residuals outweighs the error it adds. The exception is Llama on LongBench, where quantisation alone recovers only 96.3\%. In this case the additional error is not fully offset, and adding quantisation lowers recovery at every compression ratio.

\begin{table}[h!]
    \centering
    \begin{tabular}{llcccc}
        \toprule
        & & \multicolumn{2}{c}{RULER (4K)} & \multicolumn{2}{c}{LongBench} \\
        \cmidrule(lr){3-4} \cmidrule(lr){5-6}
        Model & Method & CR & Score (\%) & CR & Score (\%) \\
        \midrule
        \multirow{6}{*}{Mistral-7B}
            &  Eviction          & \greycell $2\times$  & \greycell 99.9 & \greycell $2\times$  & \greycell 99.4 \\
            \addlinespace[2pt]
            &  Quantisation       & \greycell $5.1\times$ & \greycell 99.5 & \greycell $5.1\times$        & \greycell 98.9 \\
            \addlinespace[2pt]
            &  LowRAM               & \bluecell $4\times$  & \bluecell 96.0 & \bluecell $4\times$  & \bluecell 97.9 \\
            \addlinespace[2pt]
            & \quad LowRAM-E  &  $8\times$ & 91.8 &  $8\times$ & 96.6   \\
            \addlinespace[2pt]
            & \quad LowRAM-Q        & $4.8\times$ & 99.6 & $6.4\times$  & 99.4 \\
            \addlinespace[2pt]
            & \quad LowRAM-EQ    & $6.7\times$  & 93.6  & $10.7\times$ & 96.9 \\
        \midrule
        \multirow{6}{*}{Llama-3.1-8B}
            &  Eviction          & \greycell $2\times$  & \greycell 99.6 & \greycell $2\times$  & \greycell 98.8 \\
            \addlinespace[2pt]
            &  Quantisation       & \greycell $5.1\times$ & \greycell 99.8 & \greycell $5.1\times$ & \greycell 96.3 \\
            \addlinespace[2pt]
            &  LowRAM               & \bluecell $4\times$  & \bluecell 93.9 & \bluecell $4\times$  & \bluecell 99.6 \\
            \addlinespace[2pt]
            & \quad LowRAM-E & $8\times$ & 79.9  & $8\times$ & 90.6 \\
            \addlinespace[2pt]
            & \quad LowRAM-Q & $4.8\times$ & 99.4 & $6.1\times$ & 95.8 \\
            \addlinespace[2pt]
            & \quad LowRAM-EQ    & $6.7\times$  & 96.4 &  $8.8\times$ & 91.3  \\
        \bottomrule
    \end{tabular}
    \caption{Recovered performance of LowRAM and LowRAM combined with eviction (LowRAM-E), quantisation (LowRAM-Q), and both (LowRAM-EQ). The achieved compression ratio varies across benchmarks because it depends on sequence length. CR refers to compression ratio.}
    \label{tab:stacking_results_3x}
\end{table}
Eviction introduces a clearer trade-off, since it shortens the effective sequence without reducing the fixed cost of the reconstruction network, leaving less memory for residuals. At moderate compression, stacking remains beneficial for Mistral, where LowRAM-E at 4$\times$ outperforms eviction alone on both RULER (99.2\% vs.\ 98.0\%) and LongBench (97.8\% vs.\ 96.3\%), though not for Llama. As compression increases, however, the fixed cost of the network becomes limiting: Figure~\ref{fig:memory_footprint} shows that at 4$\times$ amortisation it dominates below roughly 2-3K tokens, a regime that RULER's 4K-token sequences enter after moderate eviction. Accordingly, raising the LowRAM ratio from 2$\times$ to 4$\times$ while keeping eviction fixed at 2$\times$ reduces RULER recovery by 7.4 percentage points for Mistral and 17.4 for Llama (Table~\ref{tab:stacking_results}). LongBench is less affected, with reductions of 1.2 and 3.7 points, because although its shortest tasks (around 1K tokens) have little residual budget to begin with, its longer tasks (up to 18K tokens) remain well above this threshold even after eviction.

\begin{table}[t]
    \centering
    \small
    \setlength{\tabcolsep}{4pt}
    \renewcommand{\arraystretch}{0.95}
    \begin{tabular}{lcccc}
        \toprule
        & Eviction & Quantisation & LowRAM & LowRAM-EQ \\
        \midrule
        CR
            & \(10.0\times\)
            & \(7.5\times\)
            & \(9.0\times\)
            & \(26.9\times\) \\
        Score (\%)
            & 91.6
            & 92.3
            & 90.7
            & 94.5 \\
        \bottomrule
    \end{tabular}
    \caption{
        Maximum evaluated compression retaining at least \(90\%\) of Mistral-7B performance on LongBench contexts longer than 10K tokens. CR uses the byte-level accounting in Appendix~\ref{app:comp_ratio}.
    }
    \label{tab:longbench_10k}
\end{table}

Therefore, to evaluate extreme compression without conflating compression error with the fixed overhead of the reconstruction network, we evaluate on LongBench samples longer than 10K tokens. Table~\ref{tab:longbench_10k} reports the maximum compression each method can achieve while retaining at least 90\% of baseline performance (see Table~\ref{tab:longbench_10k_full} for a sweep over compression ratios). This subset provides a more representative setting for pushing compression further, as the network parameters are sufficiently amortised to leave a meaningful residual budget. The full stack achieves 26.9$\times$ compression while recovering 94.5\% of baseline performance, matching or exceeding the standalone methods despite substantially higher compression ratios.

\subsection{Value Cache Amortisation in Isolation}\label{sec:test_time_amortisation}
To isolate value compression from key-compression error, we keep the keys exact and compare amortisation against xKV applied to the value cache (Table~\ref{tab:value_only_best}; Llama-3.1-8B, two-layer per-head MLP, 300 updates, no warm-starting or $W_{\mathrm{lin}}$). Value-cache amortisation remains close to lossless across the evaluated compression ratios, retaining 98.2\% of RULER performance and 99.3\% of LongBench performance at $4\times$ compression. xKV edges slightly ahead on RULER at higher compression ratios, but amortisation is consistently stronger on LongBench, where it stays ahead at every compression ratio. This is consistent with our motivation for treating keys and values asymmetrically. Appendix~\ref{appdx:extended_results} (Table~\ref{tab:value_only_sweep}) reports the complete sweep over adaptation budgets and the effect of the Slim initialisation $W_{\mathrm{lin}}$. Those results show that the Slim prior substantially improves performance under short adaptation budgets, but that its advantage narrows as the predictor converges.

\begin{table}[t]
    \centering
    \small
    \setlength{\tabcolsep}{4.5pt}
    \begin{tabular}{lrrrrrr}
        \toprule
        & \multicolumn{3}{c}{RULER 4K (\%)} & \multicolumn{3}{c}{LongBench (\%)} \\
        \cmidrule(lr){2-4}
        \cmidrule(lr){5-7}
        Method & \(2\times\) & \(3\times\) & \(4\times\) & \(2\times\) & \(3\times\) & \(4\times\) \\
        \midrule
        Amortisation & 100.0 & 99.6 & 98.2 & 100.4 & 100.4 & 99.3 \\
        xKV & 99.9  & 99.7  & 99.3  & 95.0  & 95.3  & 95.8 \\
        
        \bottomrule
    \end{tabular}
    \caption{
        Recovered performance of amortisation and xKV applied to the value cache in isolation on
        Llama-3.1-8B, reported as a percentage of the uncompressed-cache
        baseline. The displayed configuration uses a two-layer per-head MLP
        with 300 per-context gradient updates, without warm-starting or
        \(W_{\mathrm{lin}}\). Table~\ref{tab:value_only_sweep} provides
        the complete optimisation-budget and initialisation sweep.
    }
    \label{tab:value_only_best}
\end{table}

\section{Discussion}
Our results suggest that keys and values should be compressed differently. Within each attention head, values are largely predictable from their keys. A small per-head predictor fitted to the context, with exact residuals kept for the tokens it reconstructs worst, stays close to lossless at $4\times$ value compression and holds up better on LongBench than low-rank compression of the same cache (Table~\ref{tab:value_only_best}). Pairing it with low-rank keys gives LowRAM, which remains close to baseline at moderate compression and combines with quantisation and eviction to reach $26.9\times$ on long contexts. 

The ablations suggest why prediction suits values. The key-to-value map appears strongly head-specific (sharing one predictor across a layer's heads costs almost 20 points, Table~\ref{tab:pem_ablation_3x}). Keys are harder to predict, and their errors are likely more harmful: a key error perturbs the attention logits and can shift attention onto the wrong tokens, whereas a value error enters only after the attention weights are fixed. This matches findings on asymmetric KV quantisation \citep{tao-etal-2025-asymkv} and may explain why reconstruction error tracks downstream performance so poorly (Figure~\ref{fig:analysis}). Residual placement also matters: at a fixed budget, error-ranked selection recovers almost all accuracy, whereas random selection loses most of it (Figure~\ref{fig:residual_selection}).

LowRAM is best suited to long contexts. The predictor's parameters are a fixed cost that is only amortised after a few thousand tokens (Figure~\ref{fig:memory_footprint}). This is why LowRAM degrades more on 4K RULER than on LongBench, and why it combines less cleanly with eviction, which shortens the sequence but leaves this cost unchanged. The choice of initialisation adds a second trade-off. Meta-learning brings prefill close to baseline speed at a small cost in accuracy, while full per-context fitting is slower but more accurate. Since both components of LowRAM depend only on the context, the slower option becomes attractive whenever a context is reused, as in document QA or shared-prefix serving, because the fitting cost is paid only once.

\section{Limitations and Future Work}
We have not evaluated long generations, which would require periodically compressing the suffix of newly generated tokens (e.g. every $N$ tokens) rather than relying on a single prefill-time fit. It is also unclear how much of the learned parametrisation persists across contexts: gains from warm-starting and Slim Attention initialisation suggest part of the key-to-value relationship is shared, while gains from test-time adaptation indicate a remaining context-specific component. Understanding this split, and whether hard-to-predict tokens are stable across contexts, could enable more efficient allocation of memory and computation.

\paragraph{AI Use Statement.}
We used generative AI to edit text we had written ourselves and to write and edit code, all of which we reviewed and verified before use. We take responsibility for the final content of this work.

\paragraph{Reproducibility Statement.}
Reproduction settings are in the appendix: value-reconstruction network (\ref{app:reconstruction_training}), eviction/quantisation (\ref{sec:experimental_setup}), meta-learning (\ref{app:meta_learning}), selective reconstruction (\ref{app:selective_reconstruction}), benchmarks (\ref{appdx:benchmarks}).

\bibliography{references}
\bibliographystyle{iclr2027_conference}

\newpage
\appendix

\section{Extended Related Works}\label{appdx:ext_related_works}

\subsection{Test-time Adaptation and Learned Associative Memory}
A separate line of work reframes sequence modelling itself as online learning at inference. The motivating observation is that the distribution of sequences encountered at test time differs substantially from the training distribution, and that a model which can fit itself to the stream it is currently processing should enjoy better in-context performance.

Test-time training \citep{pmlr-v119-sun20b} gives this idea a concrete architectural form. Rather than maintaining a fixed memory state, \citet{sun2025learninglearntesttime} treat the state itself as the parameters of a small predictive model, updated during inference by a self-supervised reconstruction loss over the incoming tokens. This makes the memory a sequence-specific map learned online, rather than a hand-specified accumulation rule or a static summary of the past. Titans~\citep{behrouz2025titans} similarly add a neural long-term memory trained at test time, using a surprise-based gradient update and adaptive forgetting mechanism, and combine it with short-window attention for longer-range storage.

This view has roots in Fast Weight Programmers \citep{6796337, schmidhuber1993reducing}, where a slow controller network writes key-value associations into fast weights. \citet{pmlr-v139-schlag21a} later showed that linear attention \citep{pmlr-v119-katharopoulos20a} can be interpreted as such a fast-weight memory, with later variants improving the write rule using delta-style corrections and gates \citep{yang2024gated, NEURIPS2024_d13a3eae,yang2025gated}. We inherit the core abstraction of this line (a map from keys to values trained by a reconstruction loss) but not its deployment regime: the map is fit once, after prefill, to the completed KV cache of a pretrained Transformer, and neither the base model nor the attention computation is modified at inference.

\section{Reconstruction Network and Optimisation Details}
\label{app:reconstruction_training}

\paragraph{Architecture.}
For each Transformer layer $\ell$ and KV head $h$, we fit an independent reconstruction network
$f_{\theta_{\ell,h}}:\mathbb{R}^{d_k}\rightarrow\mathbb{R}^{d_v}$ that maps keys to values, repeated here from Equation~\eqref{eq:predictor_arch} in Section~\ref{sec:methods} with the hidden-width factor made explicit.
The predictor is a two-layer MLP with hidden width $\gamma d_k$, where $\gamma$ is the hidden-width factor, and uses ReLU as the activation function $\sigma$:
\begin{equation}
    f_{\theta_{\ell,h}}(k)
    =
    W^{(2)}_{\ell,h}
    \sigma\!\left(
        W^{(1)}_{\ell,h}k+b^{(1)}_{\ell,h}
    \right)
    + b^{(2)}_{\ell,h}
    + W^{\mathrm{lin}}_{\ell,h}k
\end{equation}
The linear branch is initialised from the Slim Attention prior,
$W^{\mathrm{lin}}_{\ell,h}=(W^K_{\ell,h})^{+}W^V_{\ell,h}$,
while the MLP learns a non-linear correction to this mapping. Although each head has separate parameters, all heads within a layer are evaluated and optimised in parallel by storing their parameters with the head dimension as an additional batch dimension. This preserves head-specific key--value mappings without requiring a separate training loop for every head.

\begin{algorithm}[h]
\caption{Value Cache Amortisation}
\label{alg:pem}
\begin{algorithmic}[1]
\STATE Input KV cache $(K,V)$, residual budget $p$
\STATE Initialise $f_\theta$ from warm-started parameters
\STATE Train $f_\theta$ on $(K,V)$ using MSE loss
\STATE $\hat{V} \gets f_\theta(K)$
\STATE $E \gets \|\hat{V} - V\|_2^2$ \hfill (per-token error)
\STATE Select top-$p$ fraction of tokens with largest errors
\STATE Store residuals $\Delta V = V - \hat{V}$ only for selected tokens
\FOR{each decoding step}
    \STATE Reconstruct values $\tilde{V} \gets f_\theta(K)$
    \STATE Add stored residuals to selected tokens
    \STATE Compute attention using $\tilde{V}$
\ENDFOR
\end{algorithmic}
\end{algorithm}

\paragraph{Layer-wise Fitting during Prefill.}
We train the reconstruction networks sequentially as the model progresses through prefill. Once attention layer $\ell$ has produced its key and value states, we fit the reconstruction networks for that layer before continuing to layer $\ell+1$. The base-model parameters remain frozen throughout.

For a layer with $H_{\mathrm{KV}}$ KV heads, we minimise the head-averaged reconstruction objective
\begin{equation}
    \mathcal{L}_{\ell}
    =
    \frac{1}{H_{\mathrm{KV}}}
    \sum_{h=1}^{H_{\mathrm{KV}}}
    \frac{1}{T}
    \left\|
        f_{\theta_{\ell,h}}(K_{\ell,h})-V_{\ell,h}
    \right\|_F^2
\end{equation}
The sequence dimension is processed in a single batch, while the head dimension is always batched.
After optimisation, we compute the reconstruction error of every token, retain residual corrections for the highest-error tokens under the specified residual budget, and discard the dense value targets and optimiser state before processing the next layer. Consequently, dense value targets and reconstruction-network optimiser states are required for only one layer at a time, substantially reducing peak memory relative to fitting all layers simultaneously.

\paragraph{Key-cache quantisation crossover.}
xKV factorises the layer-grouped key cache as $K_l \approx A B_l$, where $B_l$ is a layer-specific reconstruction matrix and $A$ holds the left singular vectors of the shared low-rank subspace. With layer-group size $g$, this stores $A$ in $\mathcal{O}(rT)$ memory and $B$ in $\mathcal{O}(r g D_k)$, where $T$ is the context length, $r$ the decomposition rank, and $D_k$ the flattened key width of one layer. Since only $A$ scales with $T$, quantising it captures most of the memory reduction at long context. The basis $B$ is independent of $T$: quantising it would save only $\mathcal{O}(r g D_k)$ while injecting error into every reconstructed key, and with $g=4$ the crossover is roughly $T \approx 4 D_k$, beyond which quantising $B$ is not worth its cost -- this is why we quantise only $A$ in the main experiments.

\begin{table}[t]
    \centering
    \label{tab:reconstruction_hyperparameters}
    \small
    \begin{tabular}{lc}
        \toprule
        Hyperparameter & Setting \\
        \midrule
        Number of MLP layers        & 2 \\
        Hidden-width factor $\gamma$& 1 \\
        Activation                  & ReLU \\
        Loss                        & MSE \\
        Optimiser                   & Adam \\
        Learning rate               & 0.001 \\
        \bottomrule
    \end{tabular}
    \caption{
        Hyperparameters used to train the value-reconstruction
        networks used for value cache amortisation.
    }
\end{table}

\section{Compression Ratio Calculation}\label{app:comp_ratio}
\begin{equation}
    \text{CR} = \frac{\text{uncompressed key bytes + uncompressed value bytes}}{\text{compressed key bytes + compressed value bytes}}
\end{equation}

since $\text{uncompressed key bytes = uncompressed value bytes}$, we can write the compression ratio as the harmonic mean:
\begin{equation}
    \text{CR} = \frac{2}{\frac{1}{\text{r}_K}+\frac{1}{\text{r}_V}} = \frac{2\text{r}_K\text{r}_V}{\text{r}_K +\text{r}_V}
\end{equation}

\textbf{Keys} are compressed across layers. Groups of $G$ adjacent attention layers are flattened and horizontally concatenated into one token-by-feature matrix, which is factorised as $M \approx AB$ at a rank chosen to hit a target ratio. What is stored is $A$ and $B$. $b$ is the bytes per element.
\begin{equation}
    r_K = \frac{M \cdot b}{A \cdot b + B \cdot b} 
\end{equation}
where 
\begin{equation}
    M \in \mathbb{R}^{m \times n}, \qquad A \in \mathbb{R}^{m \times k}, \qquad B \in \mathbb{R}^{k \times n}
\end{equation}
so 
\begin{equation}
    r_K = \frac{m \cdot n \cdot b}{m \cdot k \cdot b + k \cdot n \cdot b} =\frac{m \cdot n}{k \cdot (m +  n)} 
\end{equation}
The rank $k$ is chosen to meet compression ratio $r_K$ by solving
\begin{equation}
    k = \frac{m \cdot n}{r_K \cdot (m+n)}
\end{equation}

\textbf{Values} are compressed within a layer. The value cache stores the MLP's parameters plus the residuals and indices for the hard to predict values. For a single layer, with $n$ values, each of size $H \cdot d$,
\begin{equation}
    r_V =\frac{N \cdot  b}{p_{\theta} \cdot b + R \cdot c_{\text{res}}}
\end{equation}
where $N = n \cdot H \cdot d$, $p_{\theta}$ is the parameter count, $c_{\text{res}} = d \cdot b + b_{\text{idx}}$ is the cost of a residual ($b_{\text{idx}}$ is the bytes per residual index), and $R$ is the number of residuals.

The number of residuals is chosen based on the target compression ratio $r_V$,
\begin{equation}
    R = \frac{(\frac{N}{r_{V}} - p_{\theta}) \cdot b}{c_{\text{res}}}
\end{equation}

\textbf{Quantisation}, when applied, reduces the storage cost of the $A$ factor for the decomposed keys and the residual for the amortised value cache. For $A$, each of the $m$ rows becomes $k \cdot q(\beta_K)$ bytes of packed indices plus one norm ,
\begin{equation}
    r_K  = \frac{m \cdot n \cdot b}{m \cdot (k\cdot q(\beta_K) + b) +  k \cdot n \cdot b} 
\end{equation}
and for the residual count,
\begin{equation}
    R = \frac{\left(\frac{N}{r_{V}} - p_{\theta}\right) \cdot b}{c_{\text{packed\_res}}} \qquad\text{with} \qquad c_{\text{packed\_res}} = \underbrace{d \cdot q(\beta_V)}_{\text{packed codes}} + \underbrace{b}_{\text{norm}} + \underbrace{b_{\mathrm{idx}}}_{\text{sparse index}}
\end{equation}

\textbf{Eviction}, when applied, index-selects the key and value states down to the retained positions before either cache sees them. For layer $\ell$, the
predictor's value-importance scores select

$$n_{\text{keep}}=\rho T$$

positions, with the first 8 (attention sinks) and the last 64 (local
window) forced in and the remainder filled by top importance. The total compression ratio becomes
\begin{equation}
    \text{CR} = \frac{1}{\rho} \cdot \frac{2\text{r}_K\text{r}_V}{\text{r}_K +\text{r}_V}
\end{equation}

\section{Extended results}\label{appdx:extended_results}
\subsection{Ablation Studies}
To better understand which factors drive the performance of the amortised value cache, we perform a series of ablation studies. We first quantify the contribution of each component of the proposed method, including residual storage, Slim initialisation, warm-starting, and per-head modelling. We then compare alternative Slim initialisation schemes to determine how best to exploit the pretrained key-value relationship.

% \paragraph{MLP Hyperparameters.} We first investigate the capacity and optimisation requirements of the value reconstruction network. Table \ref{tab:mlp_perhead_trainable_slim_init_longbench_3x} reports recovered LongBench performance at 3$\times$ compression while varying the number of gradient updates and MLP depth. Increasing the optimisation budget consistently improves performance, with recovery rising from 91.2\% at 100 updates to 98.7\% at 200 updates.

% In contrast, increasing model depth does not improve reconstruction quality. A two-layer MLP substantially outperforms both one-layer and three-layer alternatives, achieving 98.7\% recovered performance. The degradation observed with three layers suggests that the per-context key-value relationship is relatively simple, so additional depth is unnecessary and would require more gradient updates for the same performance recovery.

\paragraph{Architecture Variants.} To examine the contribution of each component in the amortised value cache to performance,  we remove or vary each individually while keeping all others fixed. The results of this ablation can be seen in Table~\ref{tab:pem_ablation_3x}. The residual store and the per-head MLP are by far the most important. Removing residual corrections drops recovered performance from 99.0\% to 64.9\% (34.1 points), confirming that a small subset of tokens contributes disproportionately to reconstruction error and needs explicit storage. Using a linear predictor instead of an MLP leads to a recovered performance of 57.5\%, a performance drop even greater than not having a residual store. The linear map is fitted by least squares, $W_h = K_h^{+}V_h$, where $K_h^{+}$ denotes the pseudoinverse of the unrotated key matrix. Values are reconstructed as $\hat{V}_h = K_hW_h$, retaining the same residual-storage mechanism and target compression budget. Per-head reconstruction is the second most important choice: replacing per-head MLPs with a shared per-layer model costs 19.5 points, suggesting different heads learn distinct key-value geometries.

\begin{table}[h!]
    \centering
    \begin{tabular}{lcc}
        \toprule
        Method & LongBench (\%) & $\Delta$ \\
        \midrule
        Amortisation & 99.0 &  \\
        \quad w/o error-residual store & 64.9 & \textcolor{red}{-34.1} \\
        % \quad w random-residual store & & \textcolor{red}{-} \\
        \quad w/o slim initialisation & 97.2 & \textcolor{red}{-1.8} \\
        \quad w/o warm start & 97.1 & \textcolor{red}{-1.9} \\
        \quad w per-layer MLP & 79.5 & \textcolor{red}{-19.5} \\
        \quad w linear predictor & 57.5 & \textcolor{red}{-41.5} \\
        
        \bottomrule
    \end{tabular}
    \caption{Ablation of the main components of the amortised value cache method for Mistral-7B at $3\times$ value cache compression. LongBench scores are reported as percentages of the uncompressed KV cache baseline, and $\Delta$ denotes the absolute change in recovered performance relative to the full amortisation method.}
    \label{tab:pem_ablation_3x}
\end{table}

\paragraph{Residual Selection.}
To test whether the choice of which tokens receive residuals matters beyond the residual budget itself, we compare four selection strategies at equal budgets on RULER NIAH-Multikey-1 (4K tokens) with Llama-3.1-8B-Instruct. We apply the full LowRAM pipeline at 4$\times$ compression with the meta-learned initialisation (2 gradient updates). This is the regime in which the predictor is least converged, and in which residual selection should therefore matter most. \emph{Random} assigns residuals to uniformly sampled tokens. \emph{MSE} is our default, which ranks tokens by reconstruction error. The \emph{+ 8 sinks} variants additionally force residuals for the first 8 tokens (attention sinks) within the same budget. All strategies share the same compressed key cache and are evaluated on the same sequences, so differences between them isolate the effect of residual selection. We report NIAH accuracy over 200 samples and, on 16 of them, the mean squared error between attention outputs computed with the compressed and the exact KV cache.

\begin{figure}[t]
    \centering
    \includegraphics[width=\linewidth]{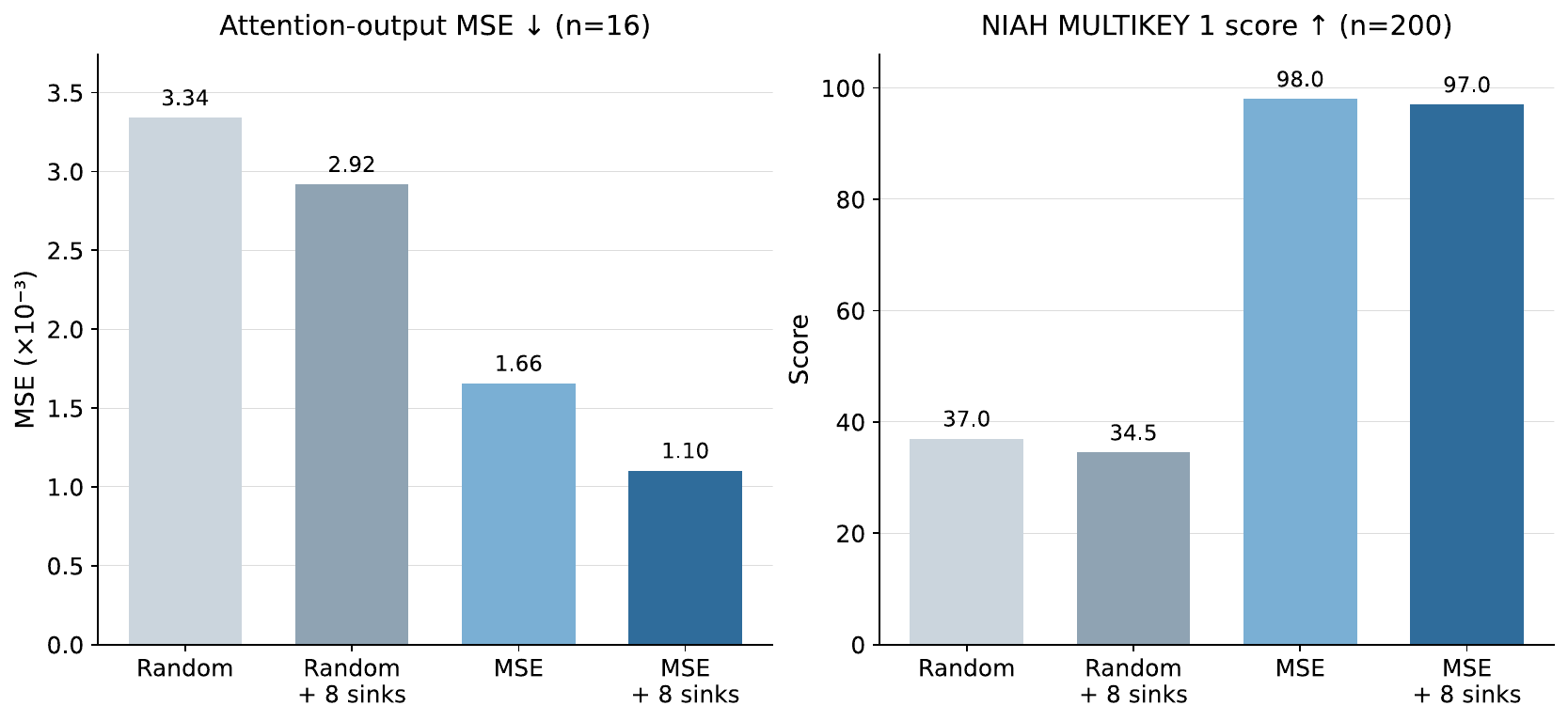}
    \caption{Residual selection strategies at equal residual budgets on RULER NIAH-Multikey-1 for Llama-3.1-8B-Instruct, using the full LowRAM pipeline at 4$\times$ compression with the meta-learned initialisation (2 gradient updates). \textbf{Left:} mean squared error between attention outputs computed with the compressed and exact KV caches ($n{=}16$). \textbf{Right:} NIAH accuracy ($n{=}200$). \emph{+ 8 sinks} forces residuals for the first 8 tokens within the same budget. All strategies share the same compressed key cache and are evaluated on the same sequences.}
    \label{fig:residual_selection}
\end{figure}

Figure~\ref{fig:residual_selection} shows that error-ranked selection is necessary. Random selection scores less than 40\%, with or without sinks, whereas MSE-ranked selection recovers 98\%. Forcing residuals for attention sinks reduces attention-output error for both random and MSE-ranked selection but does not positively affect accuracy. This observation is consistent with Section~\ref{sec:source_target} in which we show that reconstruction error is an imperfect proxy for downstream performance for values.

\paragraph{Optimisation Budget and Slim Attention Initialisation.}
We examine how the fidelity of value cache amortisation changes with the per-context optimisation budget and with the Slim Attention initialisation. These experiments isolate value compression by retaining the key cache exactly. We use Llama-3.1-8B throughout. Warm-starting is disabled so that the effect of
$W_{\mathrm{lin}}$ can be measured independently.

The plain MLP configuration learns the key-to-value map entirely from the current context. In the $+W_{\mathrm{lin}}$ configuration, the predictor additionally contains a trainable linear residual branch initialised using the Slim Attention prior,
\[
    W_{\mathrm{lin}}
    =
    W_K^{+} W_V,
\]
where \(W_K^{+}\) denotes the Moore-Penrose pseudoinverse of the key projection. The residual branch is jointly optimised with the MLP during per-context adaptation.

\begin{table}[t]
    \centering
    \small
    \setlength{\tabcolsep}{4.5pt}
    \begin{tabular}{lrrrrrrr}
        \toprule
        & & \multicolumn{3}{c}{RULER 4K (\%)} &
            \multicolumn{3}{c}{LongBench (\%)} \\
        \cmidrule(lr){3-5}
        \cmidrule(lr){6-8}
        Predictor & Updates
            & \(2\times\) & \(3\times\) & \(4\times\)
            & \(2\times\) & \(3\times\) & \(4\times\) \\
        \midrule
        MLP
            & 10
            & 90.8 & 78.0 & 61.5
            & 76.2 & 69.0 & 59.0 \\
        MLP \(+W_{\mathrm{lin}}\)
            & 10
            & 98.4 & 94.4 & 79.0
            & 84.9 & 82.3 & 78.1 \\
        \addlinespace
        MLP
            & 100
            & 99.4 & 97.6 & 88.9
            & 93.6 & 91.7 & 88.9 \\
        MLP \(+W_{\mathrm{lin}}\)
            & 100
            & 99.9 & 99.1 & 95.1
            & 94.4 & 93.8 & 91.5 \\
        \addlinespace
        MLP
            & 200
            & 99.9 & 99.2 & 97.1
            & 98.1 & 98.7 & 97.9 \\
        MLP \(+W_{\mathrm{lin}}\)
            & 200
            & 99.9 & 99.5 & 97.3
            & 98.8 & 98.4 & 97.3 \\
        \addlinespace
        MLP
            & 300
            & \textbf{100.0} & \textbf{99.6} & \textbf{98.2}
            & \textbf{100.4} & \textbf{100.4} & \textbf{99.3} \\
        \bottomrule
    \end{tabular}
    \caption{
        Recovered performance of value cache amortisation on Llama-3.1-8B as a function of the per-context optimisation budget and Slim Attention initialisation. Keys are retained exactly, and results are reported relative to the uncompressed-cache baseline. All predictors are two-layer per-head MLPs trained without warm-starting. The 300-update configuration is the high-fidelity result summarised in Table~\ref{tab:value_only_best}.
    }
    \label{tab:value_only_sweep}
\end{table}

Increasing the optimisation budget consistently improves the fidelity of the
amortised value cache. At 4$\times$ compression, the plain MLP improves from 61.5\% to 98.2\% on RULER and from 59.0\% to 99.3\% on LongBench as the budget increases from 10 to 300 updates. Performance is
largely saturated after 200-300 updates.

The Slim prior is most useful when the adaptation budget is small. With only 10 updates, $W_{\mathrm{lin}}$ improves 4$\times$ recovery by 17.5 percentage points on RULER and 19.1 points on LongBench. Its effect becomes much smaller after 200 updates, where the plain MLP has sufficient time to
learn a comparable mapping directly from the context. This indicates that $W_{\mathrm{lin}}$ primarily accelerates adaptation rather than improving the converged solution.

\paragraph{Slim Attention Initialisation.}
We consider two parameterisations. In the \emph{joint} variant, all KV heads are concatenated and a single mapping $\mathbf{W}_{\mathrm{lin}} \in \mathbb{R}^{(H_{\mathrm{KV}}d_h)\times (H_{\mathrm{KV}}d_h)}$ is estimated. This map may use key dimensions from any KV head to reconstruct the value dimensions of any other KV head. In the \emph{per-head} variant, we instead estimate one independent map per KV head, which prohibits cross-head mixing,
\begin{equation}
  \mathbf{W}_{\mathrm{lin}}^{(h)} =
  \left(\mathbf{W}_K^{(h)}\right)^{+}
  \mathbf{W}_V^{(h)}
  \in \mathbb{R}^{d_h \times d_h}
\end{equation}

For the MLP in the amortised value cache, the linear map is incorporated as a
residual branch:
\begin{equation}
  \hat{\mathbf{V}} =
  f_\theta(\mathbf{K}) + \mathbf{K}\mathbf{W}_{\mathrm{lin}}
\end{equation}
where $f_\theta$ is the learned MLP predictor. In the per-head setting,
the residual is applied independently to each head:
$\hat{\mathbf{v}}^{(h)} =
f_\theta^{(h)}(\mathbf{k}^{(h)}) +
\mathbf{k}^{(h)}\mathbf{W}_{\mathrm{lin}}^{(h)}$.
In the joint setting, the MLP remains head-wise, but its residual branch
can mix information across KV heads. The joint $W_{\mathrm{lin}}$ and per-head $W_{\mathrm{lin}}$ rows use the linear reconstruction alone, without the learned MLP correction.

Table~\ref{tab:slim_init_ablation} shows that the linear-only joint initialisation is substantially more accurate than its per-head counterpart, but requires considerably more storage. For the 8-KV-head, $d_h=128$
configuration used by both models, the joint map contains $(8 \cdot 128)^2 = 1{,}048{,}576$ parameters per layer, whereas the per-head map contains $8 \cdot 128^2 = 131{,}072$. This accounts for the much higher compression of per-head $W_{\mathrm{lin}}$ relative to joint $W_{\mathrm{lin}}$.

When $W_{\mathrm{lin}}$ is added as a residual branch and jointly optimised (10 gradient updates) with the MLP, the per-head residual consistently outperforms the joint residual. This difference may reflect both the more constrained head-wise parameterisation and the compression budget: at a fixed target compression ratio, the smaller per-head $W_{\mathrm{lin}}$ leaves more memory available for storing exact value residuals, which correct the largest reconstruction errors.

\begin{table}[t]
      \centering
      \begin{tabular}{@{}llccc@{}}
          \toprule
          Model & Value Predictor & $W_{\mathrm{lin}}$ shape & CR & LongBench (\%) \\
          \midrule
          \multirow{6}{*}{Mistral-7B}
          \vspace{0.25cm}
          & Joint $W_{\mathrm{lin}}$
              & $(H_{\mathrm{KV}},d_h,H_{\mathrm{KV}},d_h)$
              & 10$\times$ & 39.7 \\
        \vspace{0.25cm}
          & Per-head $W_{\mathrm{lin}}$
              & $(H_{\mathrm{KV}},d_h,d_h)$
              & 79$\times$ & 9.7 \\
        \vspace{0.25cm}
          & $f_\theta$ + joint residual $W_{\mathrm{lin}}$
              & $(H_{\mathrm{KV}},d_h,H_{\mathrm{KV}},d_h)$
               & 3$\times$ & 9.5 \\
          & $f_\theta$ + per-head residual $W_{\mathrm{lin}}$
              & $(H_{\mathrm{KV}},d_h,d_h)$
               & 3$\times$ & \textbf{84.7} \\
          \midrule

          \multirow{6}{*}{Llama-3.1-8B}
          \vspace{0.25cm}
          & Joint $W_{\mathrm{lin}}$
              & $(H_{\mathrm{KV}},d_h,H_{\mathrm{KV}},d_h)$
             & 11$\times$ & 67.8 \\
        \vspace{0.25cm}
          & Per-head $W_{\mathrm{lin}}$
              & $(H_{\mathrm{KV}},d_h,d_h)$
              & 89$\times$ & 20.3 \\
        \vspace{0.25cm}
          & $f_\theta$ + joint residual $W_{\mathrm{lin}}$
              & $(H_{\mathrm{KV}},d_h,H_{\mathrm{KV}},d_h)$
               & 3$\times$ & 18.5 \\
          & $f_\theta$ + per-head residual $W_{\mathrm{lin}}$
              & $(H_{\mathrm{KV}},d_h,d_h)$
              & 3$\times$ & \textbf{82.3} \\
          \bottomrule
      \end{tabular}
      \caption{Ablation of joint and per-head $W_{\mathrm{lin}}$
      initialisation for value cache amortisation on LongBench. Joint
      $W_{\mathrm{lin}}$ permits mixing across KV heads, whereas per-head
      $W_{\mathrm{lin}}$ applies an independent $d_h \times d_h$ map to each
      KV head. LongBench scores are percentages of the uncompressed baseline. CR refers to compression ratio. Differences in achieved compression ratio between Mistral and Llama reflect model-specific effective context lengths induced by their different tokenizers.}
      \label{tab:slim_init_ablation}
  \end{table}

\subsection{Sweep over Compression Ratios}
Table~\ref{tab:stacking_results} provides the full compression-performance sweep underlying Table~\ref{tab:stacking_results_3x}. It reports results for LowRAM, eviction, and quantisation, both individually and in combination, across models and benchmarks. Table~\ref{tab:longbench_10k_full} similarly expands Table~\ref{tab:longbench_10k}, showing how performance varies across compression ratios on the subset of LongBench examples exceeding 10K tokens.

\begin{table}[h!]
    \centering
    \small
    \begin{tabular}{llcccc}
        \toprule
        & & \multicolumn{2}{c}{RULER (4K)} & \multicolumn{2}{c}{LongBench} \\
        \cmidrule(lr){3-4} \cmidrule(lr){5-6}
        Model & Method & CR & \% & CR & \% \\
        \midrule
        \multirow{18}{*}{Mistral-7B}
            & \multirow{3}{*}{Eviction} & \greycell $2\times$ & \greycell 99.9 & \greycell $2\times$ & \greycell 99.4 \\
            &  & \greycell $3\times$ & \greycell 99.3  & \greycell $3\times$ & \greycell 97.6 \\
            &  & \greycell $4\times$ & \greycell 98.0 & \greycell $4\times$ & \greycell 96.3 \\
            \addlinespace[2pt]
            \cmidrule(lr){2-6}
            &  Quantisation & \greycell $5.1\times$ & \greycell 99.5 & \greycell $5.1\times$ & \greycell 98.9 \\
            \addlinespace[2pt]
            \cmidrule(lr){2-6}
            & \multirow{3}{*}{LowRAM} & \bluecell $2\times$ & \bluecell 99.4 & \bluecell $2\times$ & \bluecell 100.7 \\
            &  & \bluecell $3\times$ & \bluecell 98.8 & \bluecell $3\times$ & \bluecell 99.4 \\
            &  & \bluecell $4\times$ & \bluecell 96.0 & \bluecell $4\times$ & \bluecell 97.9 \\
            \addlinespace[2pt]
            \cmidrule(lr){2-6}
            & \multirow{3}{*}{LowRAM-E} & $4\times$ & 99.2 & $4\times$ & 97.8 \\
            & & $6\times$ & 97.3 & $6\times$ & 97.5 \\
            & & $8\times$ & 91.8 & $8\times$ & 96.6 \\
            \addlinespace[2pt]
            \cmidrule(lr){2-6}
            & \multirow{3}{*}{LowRAM-Q} & $3.3\times$ & 99.8 & $4.5\times$ & 99.9 \\
            & & $4.1\times$ & 99.6 & $5.6\times$ & 99.5 \\
            & & $4.8\times$ & 99.6 & $6.4\times$ & 99.4 \\
            \addlinespace[2pt]
            \cmidrule(lr){2-6}
            & \multirow{3}{*}{LowRAM-EQ} & $5.13\times$ & 99.1 & $7.5\times$ & 97.7 \\
            & & $6.7\times$ & 93.6 & $9.3\times$ & 97.4 \\
            & & $8.7\times$ & 89.1 & $10.7\times$ & 96.9 \\
        \midrule
        \multirow{18}{*}{Llama-3.1-8B}
            & \multirow{3}{*}{Eviction} & \greycell $2\times$ & \greycell 99.6 & \greycell $2\times$ & \greycell 98.8 \\
            &  & \greycell $3\times$ & \greycell 98.8 & \greycell $3\times$ & \greycell 96.6 \\
            &  & \greycell $4\times$ & \greycell 98.1 & \greycell $4\times$ & \greycell 95.4 \\
            \addlinespace[2pt]
            \cmidrule(lr){2-6}
            & Quantisation & \greycell $5.1\times$ & \greycell 99.8 & \greycell $5.1\times$ & \greycell 96.3 \\
            \addlinespace[2pt]
            \cmidrule(lr){2-6}
            & \multirow{3}{*}{LowRAM} & \bluecell $2\times$ & \bluecell 99.6 & \bluecell $2\times$ & \bluecell 98.7 \\
            &  & \bluecell $3\times$ & \bluecell 97.9 & \bluecell $3\times$ & \bluecell 98.5 \\
            &  & \bluecell $4\times$ & \bluecell 93.9 & \bluecell $4\times$ & \bluecell 99.6 \\
            \addlinespace[2pt]
            \cmidrule(lr){2-6}
            & \multirow{3}{*}{LowRAM-E} & $4\times$ & 97.3 & $4\times$ & 94.3 \\
            & & $6\times$ & 90.0 & $6\times$ & 92.4 \\
            & & $8\times$ & 79.9 & $8\times$ & 90.6 \\
            \addlinespace[2pt]
            \cmidrule(lr){2-6}
            & \multirow{3}{*}{LowRAM-Q} & $3.4\times$ & 99.9 & $4.4\times$ & 96.5 \\
            & & $4.2\times$ & 99.7 & $5.3\times$ & 95.8 \\
            & & $4.8\times$ & 99.4 & $6.1\times$ & 95.8 \\
            \addlinespace[2pt]
            \cmidrule(lr){2-6}
            & \multirow{3}{*}{LowRAM-EQ} & $5.2\times$ & 98.2 & $7.1\times$ & 93.5 \\
            & & $6.7\times$ & 96.4 & $8.8\times$ & 91.3 \\
            & & $8.8\times$ & 82.7 & $10.2\times$ & 89.4 \\
        \bottomrule
    \end{tabular}
    \caption{Recovered performance of LowRAM, used alone or combined with eviction and quantisation, across a range of compression settings for Mistral-7B and Llama-3.1-8B. Results are reported as percentages of the corresponding uncompressed baseline on RULER 4K and LongBench. Grey rows show eviction and quantisation applied independently, blue rows show LowRAM alone, and unshaded rows show LowRAM combined with eviction (LowRAM-E), quantisation (LowRAM-Q), or both (LowRAM-EQ). CR denotes the total compression ratio achieved on each benchmark; for combined methods, it accounts for the composition of the individual compression components and may differ between benchmarks because of sequence-length variation and across models due to model-specific effective context lengths induced by their different tokenizers.}
    \label{tab:stacking_results}
\end{table}

\begin{table}[t]
    \centering
    \begin{tabular}{lcc}
        \toprule
        & \multicolumn{2}{c}{LongBench ($>$10K)} \\
        \cmidrule(lr){2-3}
        Method & CR & Score (\%) \\
        \midrule

        \multirow{4}{*}{Eviction}
            & $3.0\times$ & 98.1 \\
            & $4.0\times$ & 97.8\\
            & $8.0\times$ & 93.9 \\
            & $10.0\times$ &  91.6\\
        \addlinespace[2pt]
        \cmidrule(lr){1-3}

        \multirow{4}{*}{Quantisation}
            & $3.9\times$ & 100.0\\
            & $5.1\times$ & 99.3\\
            & $7.5\times$ & 92.3 \\
            & $14.2\times$ & 33.4\\
        \addlinespace[2pt]
        \cmidrule(lr){1-3}

        \multirow{6}{*}{LowRAM}
            & $3.0\times$ & 99.6\\
            & $4.0\times$ & 97.2 \\
            & $6.0\times$ & 95.2\\
            & $7.0\times$ & 93.4 \\
            & $8.0\times$ & 91.9 \\
            & $9.0\times$ & 90.7 \\
            & $10.0\times$ &  89.0 \\
        \addlinespace[2pt]
        \cmidrule(lr){1-3}

        \multirow{4}{*}{LowRAM-EQ}
            & $9.2\times$  & 99.2 \\
            & $17.8\times$ & 98.0 \\
            & $22.1\times$ & 96.2 \\
            & $26.9\times$ & 94.5 \\

        % \addlinespace[2pt]
        % \cmidrule(lr){1-3}
        % \multirow{3}{*}{Eviction + Quantisation}
        %     & $9.2\times$  & - \\
        %     & $15.7\times$  & - \\
        %     & $22.1\times$ & - \\

        \bottomrule
    \end{tabular}

    \caption{Compression ratio and recovered performance for Mistral-7B using different KV cache compression methods, evaluated on the subset of LongBench examples with sequence lengths exceeding 10k tokens. Scores are reported as percentages of the uncompressed baseline. CR refers to compression ratio and is calculated from the byte-level key and value footprints described in Appendix~\ref{app:comp_ratio}}

    \label{tab:longbench_10k_full}
\end{table}

\section{Meta-learning}
\label{app:meta_learning}
We reduce the cost of fitting the value cache MLP by meta-learning an initialisation from which the reconstruction network can rapidly memorise the key-to-value mapping of a previously unseen sequence. This differs from the generalisation-after-adaptation objective typically considered in meta-learning. During prefill, all key-value pairs are observed, and the adapted predictor is evaluated on its ability to reconstruct those same values. Our objective is therefore rapid fitting of the observed context rather than generalisation to a held-out query set.

We use First-Order Model-Agnostic Meta-Learning (FOMAML). Let
\begin{equation}
    S = \left\{(k_i, v_i)\right\}_{i=1}^{T}
\end{equation}

denote one meta-learning task, consisting of the key-value pairs from a single sequence. We omit layer and head indices for clarity; the same procedure is applied to each per-head reconstruction network. The reconstruction loss for $S$ is
\begin{equation} 
    \mathcal{L}_{S}(\theta) = \frac{1}{T} \sum_{i=1}^{T} \left\lVert f_{\theta}(k_i)-v_i \right\rVert_{2}^{2}
\end{equation}

Starting from the shared initialisation $\phi$, the task-specific parameters are adapted for $K$ inner-loop updates:
\begin{align} 
    \theta_{S}^{(0)} &= \phi, \\ \theta_{S}^{(j+1)} &= \mathcal{U}_{\alpha} \left( \theta_{S}^{(j)}, \nabla_{\theta_{S}^{(j)}} \mathcal{L}_{S}\!\left(\theta_{S}^{(j)}\right) \right), \qquad j = 0,\ldots,K-1
\end{align} 

where $\mathcal{U}_{\alpha}$ denotes one update of the inner optimiser with learning rate $\alpha$. The same sequence is used in the inner and outer loops because the desired behaviour is rapid memorisation of the observed key-value pairs.

The meta-loss is evaluated only after the final inner-loop update:
\begin{equation}
\mathcal{L}_{{\mathrm{meta}}}(S;\phi)
=
\mathcal{L}_{S}\left(\theta_{S}^{(K)}\right)
\end{equation}

Following FOMAML, we ignore derivatives through the inner-loop optimisation trajectory. For a sequence $S_b$, its first-order contribution to the meta-gradient is therefore
\begin{equation}
g_{b}^{\mathrm{meta}}
=
\nabla_{\theta_{S_b}^{(K)}}
\mathcal{L}_{S_b}\left(\theta_{S_b}^{(K)}\right)
\end{equation}

This avoids the second-order derivatives that would otherwise arise when differentiating the final adapted parameters $\theta_{S_b}^{(K)}$ with respect to $\phi$.

During meta-training, we process one sequence at a time. For each sequence $S_b$, the reconstruction network is independently initialised from the current shared parameters $\phi$ and adapted for $K=2$ inner-loop updates. We compute its final-step meta-gradient $g_b^{\mathrm{meta}}$ and accumulate this gradient without performing an outer update. After processing $N$ sequences, we average the accumulated gradients:
\begin{equation}
\bar{g}^{\mathrm{meta}} = \frac{1}{N} \sum_{b=1}^{N} g_b^{\mathrm{meta}}
\end{equation}
and use $\bar{g}^{\mathrm{meta}}$ to perform one outer-optimiser update of $\phi$. Because no outer update occurs during the accumulation window, all $N$ sequences begin their inner-loop adaptation from the same current initialisation. The effective meta-batch size is therefore $N$ sequences, although only one sequence and its adapted reconstruction network are processed at a time.

At inference time, each reconstruction network is initialised from the learned parameters $\phi^{\star}$ and adapted to the new sequence for $K=2$ updates using its complete set of key-value pairs. Meta-training is performed offline and therefore does not contribute to TTFT.

\begin{table}[t]
    \centering 
    \small 
    \begin{tabular}{llc} 
    \toprule 
    Component & Hyperparameter & Setting \\ 
    \midrule 
    Meta-optimisation & Algorithm & FOMAML \\ 
    & Inner-loop updates, $K$ & 2 \\ 
    & Inner optimiser & Adam \\ 
    & Inner learning rate, $\alpha$ & $10^{-3}$ \\ 
    & Outer optimiser & AdamW \\ 
    & Outer learning rate, $\beta$ & $3 \times 10^{-4}$ \\ 
    & Outer-loop gradient accumulation & 8 \\ 
    & Inner-loop batch size & 4 \\ 
    \midrule 
    Predictor & Number of layers & 2 \\
    & Hidden-width factor, $\gamma$ & 1 \\ 
    & Activation & ReLU \\ & Objective & MSE \\ 
    \bottomrule 
    \end{tabular} 
\caption{Hyperparameters used to meta-learn the value-predictor initialisation. The inner and outer objectives use the same key-value pairs because the desired behaviour is rapid memorisation of the observed context.} \label{tab:meta_learning_hparams} 
\end{table}

\begin{table}[h!]
    \centering
    \small
    \begin{tabular}{cccccc}
        \toprule
            \multicolumn{3}{c}{RULER 4K (\%)} & \multicolumn{3}{c}{LongBench (\%)} \\
            \cmidrule(lr){1-3} \cmidrule(lr){4-6}
            2$\times$ & 3$\times$ & 4$\times$ & 2$\times$ & 3$\times$ & 4$\times$ \\
        \midrule
            98.4 \dd{1.2} & 96.1 \dd{1.8} & 91.2 \dd{2.7} & 93.6 \dd{5.1} & 93.0 \dd{5.5} & 91.1 \dd{8.5} \\
        \bottomrule
    \end{tabular}
    \caption{Recovered performance of LowRAM when the amortised value cache is initialised from meta-learned weights and adapted with only 2 gradient steps for Llama-3.1-8B. Scores are reported as percentages of the uncompressed baseline; values in parentheses give the drop in percentage points relative to vanilla LowRAM (Table~\ref{tab:v_only_decomp_results}), which uses 150 gradient steps per context.}
    \label{tab:meta_learning_lowram}
\end{table}

\section{Selective Reconstruction}
\label{app:selective_reconstruction}

% \paragraph{Selective reconstruction.} At $4\times$ compression, selective reconstruction (Section~\ref{sec:efficiency_methods}) preserves accuracy, $96.7$-$98.6\%$ of baseline RULER performance across 16K-64K contexts (Table~\ref{tab:quality_selective_reconstruction}), while substantially improving decode throughput (2.2$\times$ at 262K context) and attention latency at long context and large batch size, where the uncompressed baseline becomes memory-bound or OOMs entirely (Figure~\ref{fig:throughput_main}; Appendix~\ref{app:selective_reconstruction}, Figure~\ref{fig:latency}, gives the attention-latency counterpart).

To reduce decode-time compute, we adopt landmark-guided selective reconstruction, following ShadowKV~\citep{sun2025shadowkv}: rather than reconstructing the full key/value cache at every decode step, we retrieve and reconstruct only a query-dependent subset of tokens, using inexpensive per-chunk landmarks as a cheap proxy for full attention when deciding which tokens to retrieve.

\paragraph{Prefill: landmark construction and exact retention.}
During prefill, keys are partitioned into fixed-size chunks of $c$ tokens, and a landmark is computed for each chunk as the mean of its keys. Two subsets of chunks are then set aside and kept exact (both keys and values, uncompressed) rather than represented by their landmark: the $o$ chunks whose keys are least similar to their own landmark (outlier chunks, as in ShadowKV), and the chunks overlapping the most recent $w$ tokens (a local window; unlike the landmark and outlier mechanism, this is not part of ShadowKV's original design). The remaining chunks form the landmark pool used for query-time retrieval at decode time.

This exact retention is not free: the landmarks, their chunk indices, and the outlier/local exact keys and values are all additional persistent state on top of the compressed key cache itself. Rather than reporting a compression ratio that ignores this, we account for it directly in rank selection. Because xKV shares one low-rank factor across every layer in a group rather than factorising each layer independently, this overhead is measured per layer but spent per group: at the point a group's decomposition is triggered, we sum the selective-reconstruction overhead of every layer in the group and subtract it from the byte budget implied by the group's target compression ratio before choosing the shared rank, so the group's realised ratio (low-rank factors plus the combined overhead of its member layers) matches the configured target.

We additionally accelerate the low-rank key decomposition using a custom randomised SVD kernel adapted from xKV~\citep[Appendix~C.2]{chang2026xkv}, which executes power iterations in 16-bit precision and replaces Householder QR with a parallelisable shifted Cholesky-QR factorisation, in place of exact SVD when factorising the key matrices. This modification leads to a 4$\times$ speedup over exact SVD without harming performance.
%(see Table~\ref{tab:quality_selective_reconstruction}).

\paragraph{Decoding: landmark-guided retrieval.}
Landmarks are computed per key/value head, so under grouped-query attention we pool scores across each group of query heads sharing a head by taking the max, before selecting chunks. At each decode step, the incoming query is scored against every landmark in the pool, and the top-$k$ chunks (a chunk budget $k$, equal to the token budget in Table~\ref{tab:selective_reconstruction_config} divided by chunk size $c$) are retrieved and expanded back into token positions. The retrieved positions are combined with the exact outlier and local positions; keys and values, though compressed independently, are reconstructed at this shared set of positions.

\paragraph{Persistent reconstruction cache.}
Following ShadowKV, we avoid redundant reconstruction work across decode steps by keeping a persistent buffer of the $k$ most recently reconstructed chunks (capacity $k \times c$ tokens). At each step, the newly selected chunk set is diffed against what the buffer already holds: chunks selected again are reused as-is, and only chunks newly entering the selection are reconstructed and written into the buffer. This is implemented via fused CUDA kernels adapted from ShadowKV/xKV.

\begin{table}[t]
    \centering
    \begin{tabular}{lc}
        \toprule
        Parameter & Value \\
        \midrule
        Chunk size & 8 tokens \\
        Token budget & 2048 tokens \\
        Local window & 32 tokens \\
        Outlier chunks & 48 chunks (384 tokens) \\
        \midrule
        Max.\ tokens attended per step & 2464 tokens \\
        \bottomrule
    \end{tabular}
    \caption{Selective reconstruction configuration.}
    \label{tab:selective_reconstruction_config}
\end{table}

Figure~\ref{fig:throughput_main} presents the results for decode throughput, comparing LowRAM-SR (LowRAM with landmark-guided selective retrieval) at 4$\times$ compression against the uncompressed baseline at the same context length and batch size. 

% To assess whether selective retrieval affects downstream task quality, we evaluate the compressed cache on RULER at different sequence lengths. We additionally compare the two low-rank factorisation backends available for the key cache: an exact SVD, and a randomised Cholesky-QR-based approximation. The quality results can be seen in Table~\ref{tab:quality_selective_reconstruction}.

% \begin{table}[t]
%     \centering
%     \begin{tabular}{lcccc}
%         \toprule
%         & & \multicolumn{3}{c}{RULER (\%)} \\
%         \cmidrule(lr){3-5}
%         SVD backend & CR & 16K & 32K & 64K \\
%         \midrule
%         Exact SVD    & \(4\times\) & 98.1 & 98.3 & 96.7 \\
%         Cholesky--QR & \(4\times\) & 98.1 & 98.6 & 97.3 \\
%         \bottomrule
%     \end{tabular}
%     \caption{Recovered RULER performance at 4$\times$ compression under selective reconstruction (Table~\ref{tab:selective_reconstruction_config} configuration), relative to the uncompressed baseline, at increasing context lengths for Llama-3.1-8B-Instruct and for different SVD backends.}
%     \label{tab:quality_selective_reconstruction}
% \end{table}

\begin{figure}
    \centering
    \includegraphics[width=0.7\linewidth]{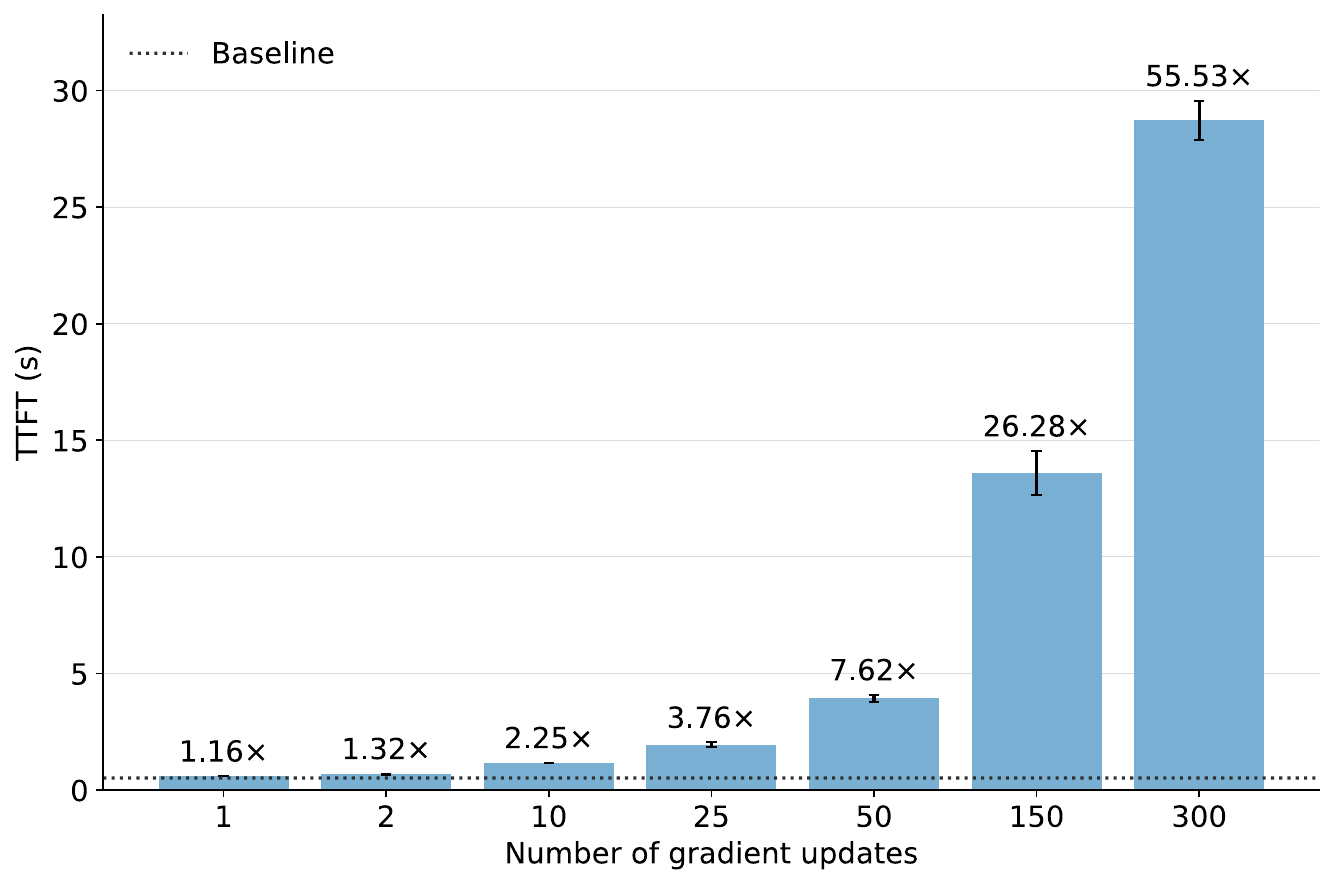}
    \caption{TTFT overhead of value cache amortisation alone versus the number of gradient updates used to fit the MLP, at a 4K-token sequence length with batch size 4. Keys are left uncompressed, so this isolates the cost of fitting the value predictor. Bars show raw TTFT; annotations give the overhead relative to the uncompressed baseline. See Figure~\ref{fig:ttft_sweep} for the full LowRAM pipeline, including key factorisation.}
    \label{fig:ttft_vs_epochs}
\end{figure}

\begin{figure}
    \centering
    \includegraphics[width=1.0\linewidth]{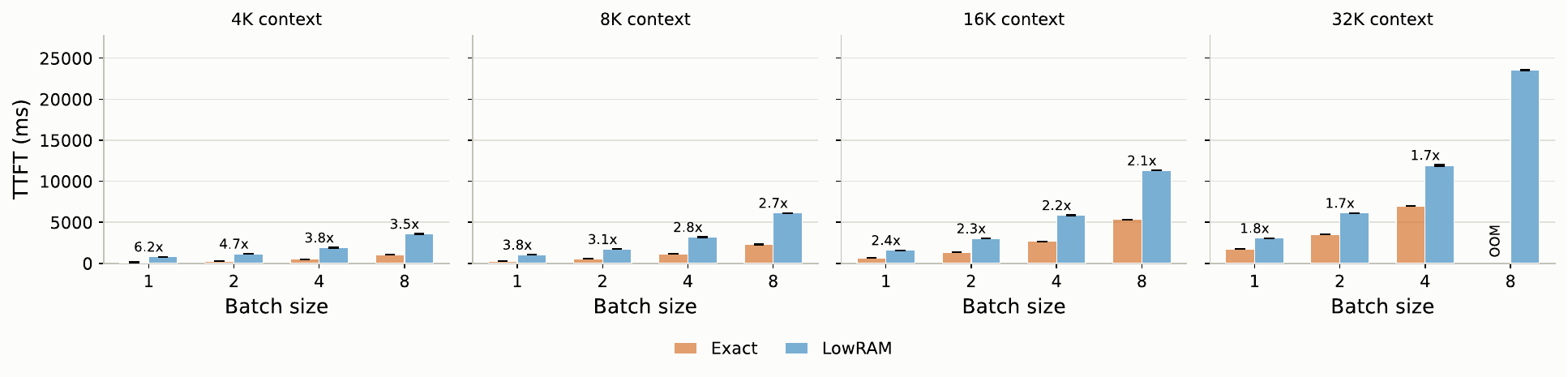}
    \caption{TTFT of the uncompressed baseline (exact KV) and the full LowRAM pipeline at 4$\times$ compression across context lengths and batch sizes. LowRAM timings include both value cache amortisation (2 gradient updates) and low-rank factorisation of the keys via randomised Cholesky-QR, so overheads exceed those of the value cache alone (Figure~\ref{fig:ttft_vs_epochs}). Annotations give the overhead relative to the baseline; OOM marks configurations where the baseline runs out of memory.}
    \label{fig:ttft_sweep}
\end{figure}

\begin{figure}
    \centering
    \includegraphics[width=1.0\linewidth]{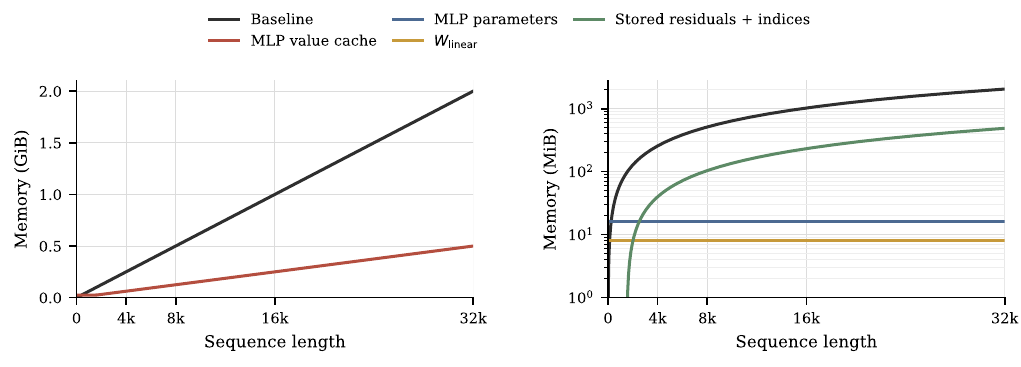}
    \caption{Analytical memory footprint of the amortised value cache as a function of sequence length. \textbf{Left:} total value cache memory for the uncompressed baseline against value cache amortisation at 4$\times$ compression. \textbf{Right:} decomposition of our footprint on a logarithmic scale. The MLP parameters and the linear residual $W_{\text{lin}}$ are constant in sequence length, while only the stored residuals and their indices scale with $T$. The fixed parameter cost dominates only at short contexts (below ${\sim}2$--$3$k tokens), and is quickly amortised as the context grows.}
    \label{fig:memory_footprint}
\end{figure}

\section{Benchmarks}\label{appdx:benchmarks}

\begin{table*}[t]
\centering
\small
\setlength{\tabcolsep}{3pt} % default is 6pt

\begin{adjustbox}{max width=\textwidth}
\begin{tabular}{@{}llrllrrr@{}}
\toprule
\multirow{2}{*}{\textbf{Dataset}} & \multirow{2}{*}{\textbf{Source}} & \multirow{2}{*}{\textbf{Avg len}} & \multirow{2}{*}{\textbf{Metric}} & \multirow{2}{*}{\textbf{Language}} & \multirow{2}{*}{\textbf{\#data}} & \multicolumn{2}{c}{\textbf{\#data ($>$10K)}} \\
\cmidrule(lr){7-8}
 & & & & & & \textbf{Mistral} & \textbf{Llama} \\
\midrule

\multicolumn{8}{l}{\textit{Single-Document QA}} \\
NarrativeQA      & Literature, Film   & 18,409 & F1             & English & 200 & 181 & 181 \\
Qasper           & Science            & 3,619  & F1             & English & 200 & 11 (excl.) & 7 (excl.) \\
MultiFieldQA-en  & Multi-field        & 4,559  & F1             & English & 150 & 46  & 24 \\
\midrule

\multicolumn{8}{l}{\textit{Multi-Document QA}} \\
HotpotQA         & Wikipedia          & 9,151  & F1             & English & 200 & 171 & 158 \\
2WikiMultihopQA  & Wikipedia          & 4,887  & F1             & English & 200 & 49  & 38 \\
MuSiQue          & Wikipedia          & 11,214 & F1             & English & 200 & 197 & 195 \\
\midrule

\multicolumn{8}{l}{\textit{Summarization}} \\
GovReport        & Government report  & 8,734  & Rouge-L        & English & 200 & 99  & 88 \\
QMSum            & Meeting            & 10,614 & Rouge-L        & English & 200 & 160 & 153 \\
MultiNews        & News               & 2,113  & Rouge-L        & English & 200 & 5 (excl.) & 4 (excl.) \\
\midrule

\multicolumn{8}{l}{\textit{Few-shot Learning}} \\
TREC             & Web question       & 5,177  & Accuracy (CLS) & English & 200 & 59  & 36 \\
TriviaQA         & Wikipedia, Web     & 8,209  & F1             & English & 200 & 136 & 123 \\
SAMSum           & Dialogue           & 6,258  & Rouge-L        & English & 200 & 111 & 90 \\
\midrule

\multicolumn{8}{l}{\textit{Synthetic Task}} \\
PassageRetrieval-en & Wikipedia       & 9,289  & Accuracy (EM)  & English & 200 & 200 & 200 \\
\midrule

\multicolumn{8}{l}{\textit{Code Completion}} \\
LCC              & Github             & 1,235  & Edit Sim       & Python/C\#/Java & 500 & 26  & 14 (excl.)\\
RepoBench-P      & Github repository  & 4,206  & Edit Sim       & Python/Java     & 500 & 307 & 229 \\
\midrule
\textbf{Total}   &                    &        &                &         & 3,550 & 1,742 & 1,515 \\
\bottomrule
\end{tabular}
\end{adjustbox}

\caption{LongBench datasets used for evaluation. \#data ($>$10K) gives the number of samples in the long-context subset used in Tables~\ref{tab:longbench_10k} and~\ref{tab:longbench_10k_full}, i.e.\ samples whose prompt exceeds 10K tokens under the Mistral-7B and Llama tokenizers, respectively. Qasper and MultiNews (and LCC for Llama) are excluded from this subset because less than 20 of their samples meet the length requirement. Avg len is reported in words, following the original LongBench statistics.}
\label{tab:longbench_datasets}
\end{table*}

\end{document}